\documentclass[designerfonts=false]{allpurpose}

\renewcommand{\eqref}[1]{equation~\ref{#1}}

\usepackage{url}
\usepackage{enumitem}
\usepackage{booktabs,tabularx}
\usepackage{colortbl}
\usepackage{multirow}
\usepackage{capt-of}
\usepackage{graphicx}
\usepackage{algorithm}
\usepackage{algpseudocode}
\usepackage{float}
\usepackage{placeins}
\usepackage{xspace}
\usepackage{tikz}
\usetikzlibrary{arrows.meta}
\usepackage[most]{tcolorbox}
\usepackage{fvextra}
\usepackage{hyperref}
\usepackage{caption}
\usepackage{hyperref}
\renewcommand{\textsc}[1]{\textup{\MakeUppercase{#1}}}
\AtBeginDocument{}

\usepackage{newunicodechar}
\newunicodechar{≥}{\ensuremath{\geq}}

\newcommand{\method}{%
  \texorpdfstring{\textrm{\textup{Meta\textsuperscript{2}}}}{Meta2}\xspace%
}

\newcommand{\appendixoutlineentry}[1]{%
  \par\addvspace{5pt}%
  \noindent
  \makebox[2em][l]{\textbf{\ref{#1}}}%
  \hyperref[#1]{\textbf{\nameref*{#1}}}%
  \dotfill
  \textbf{\pageref{#1}}\par
}

\newcommand{\appendixoutlinesubentry}[1]{%
  \par\addvspace{2pt}%
  \noindent
  \hspace*{1.5em}%
  \makebox[3em][l]{\ref{#1}}%
  \hyperref[#1]{\nameref*{#1}}%
  \dotfill
  \pageref{#1}\par
}

\title[\method: Reusable knowledge for harness evolution]{\method: Precomputing Reusable Knowledge for Online Harness Evolution in Long-Horizon Agents}

\paperstatus{\textcolor{APAccent}{\affmark{*}}Equal contribution. Preprint. Correspondence: \href{mailto:lingzhiy@umd.edu,furongh@umd.edu}
{\texttt{\{lingzhiy,furongh\}@umd.edu}}.}
\paperlink[home]{Project}{https://meta-square.github.io/}
\author{%
  Lingzhi Yuan\affmark{1,*}, Yiguang Cai\affmark{2,*}, Weize Liu\affmark{1},
  Siqi Yang\affmark{3}, Feijiang Han\affmark{1}, \mbox{\apauthor{Yongkyun Lee}},
  \mbox{Udari Madhushani Sehwag\affmark{4}}, \apauthor[1]{Furong Huang}}
\affiliations{%
  \affmark{1}University of Maryland; \affmark{2}University of Washington;
  \affmark{3}University of Illinois at Urbana-Champaign; \affmark{4}Scale AI; \apaffiliation}
\abstracttext{Harness optimization improves long-horizon agents by refining the prompts, tools, memory, and hooks around a frozen model. 
Offline methods can learn from completed trajectories, but deploy one shared harness for the task agent that may overfits training tasks and cannot react to evidence revealed during a new run. 
Online evolution can adapt before, during or after one run through an evolver, but may have to reason about every intervention from scratch, or rely on verifier feedback which is usually unavailable at test-time. 
We introduce \method, 
which moves offline optimization one level up: rather than optimizing a shared harness, it learns reusable knowledge for harness evolution and applies it within each task online.
\textbf{Offline}, \method learns a textual Meta-Skill $\mathcal{M}$ by contrasting training rollouts of the same task agent under three configurations: without an evolver, with an evolver guided by the current version of $\mathcal{M}$, and with an evolver given privileged verifier information available only during training. 
Contrasts in trajectories and outcomes are aggregated across tasks to refine $\mathcal{M}$ into reusable guidance on when and how to evolve, when to leave the harness unchanged, and which interventions to avoid.
\textbf{Online}, an evolver combines the frozen $\mathcal{M}$ with emerging trajectory evidence to decide whether to \textsc{Keep} the harness unchanged or \textsc{Evolve} it through a task-local edit. 
Each edit shapes the remainder of the same run, preserving its environment state and accumulated progress. 
This way, evolution guidance transfers across tasks while realized harness edits remain task-local, and held-out verifier feedback is never available to the evolver.
On two long-horizon benchmarks Terminal-Bench 2.1 and DeepSWE 1.1, \method achieves the highest performance among evaluated methods with GPT-5.6-Terra, reaching 81.00\% and 44.44\%, respectively. Ablations support the value of contrastive evolution guidance and keeping online harness edits task-local.
}

\begin{document}

\maketitle

\section{Introduction}
\label{sec:introduction}

Long-horizon agents do not act through a model alone. 
They act through a \emph{harness}: prompts that frame decisions, a context policy that preserves information, tools and hooks that enable actions, and checks for completion~\citep{ning2026code, lopopolo2026harness, weng2026harness, young2025effective}.
These mechanisms are often fixed before execution, yet the need to adapt them emerges during execution. 
A missing tool becomes apparent when the agent tries to use it; an overlooked constraint may surface in an intermediate artifact; and a one-time warning may need to become a recurring check. 
A harness chosen in advance must anticipate such situations before the trajectory that exposes them exists.

Because different tasks expose different interface and control needs, hand-designed harnesses~\citep{pi_coding_agent, openai2025codex, yang2024swe, deepseek-harness2026} must either anticipate a wide range of situations or be specialized manually. 
Automatic harness optimization reduces this burden, but most methods still make their consequential choices outside the active run.
Offline methods optimize prompts, memory, tools, workflows, or complete harness programs on training tasks and deploy one shared artifact~\citep{lee2026meta, lin2026agentic, zhang2026self, yang2026one, yuan2026flowbank, yuksekgonul2024textgrad, agrawal2026gepa, zhang2025aflow}. 
This shared artifact may overfit training tasks and cannot react to evidence revealed during a new run.
Online methods adapt the harness after the target task becomes available.
Some customize it from the target instruction before execution (\emph{pre-run}) or revise it using feedback from completed attempts (\emph{after-run})~\citep{zhang2026jit,shao2026harness, wang2025scoreflow}.
Recent systems support \emph{within-run} evolution by updating prompts, tools, skills, or memory during execution~\citep{karten2026continual,fu2026chill,xiao2026pilot,xia2025live}.
Such within-run harness evolution can be viewed as a practical step toward recursive self-improvement (RSI), where an agent improves not only its task behavior but also the system that shapes its subsequent behavior.
Pre-run and after-run adaptation cannot apply an edit to the unfinished run in which the motivating evidence arises.
Within-run evolution can respond to such evidence as it emerges, but the evolver still needs guidance on when and how to intervene.
This raises our central design question:
\textit{\textbf{how can lessons learned from completed training runs guide selective edits during a new run, without accessing held-out verifier feedback or making one task's correction persist into unrelated tasks?}}

\begin{figure}[t]
  \centering
  \includegraphics[width=\linewidth]{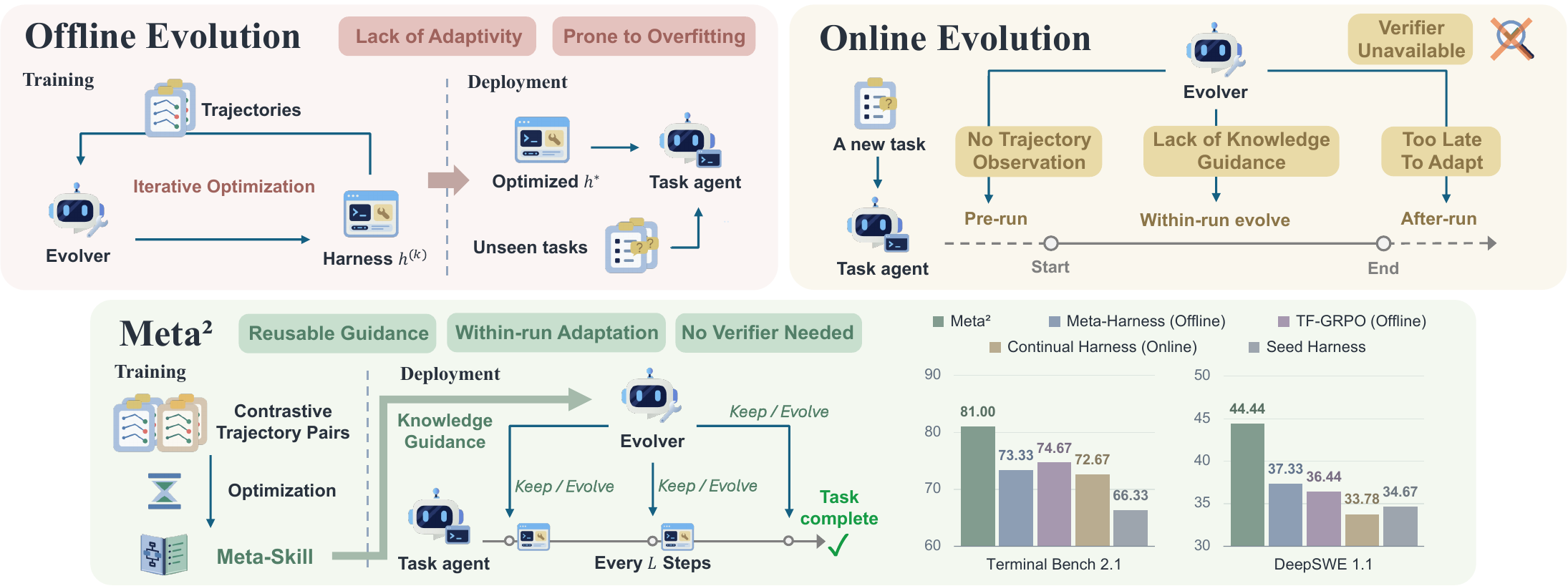}
  \vspace{-10pt}
  \caption{\textbf{Motivation and overview of \method.} 
    Offline evolution produces a shared harness that may overfit training tasks and cannot adapt to evidence revealed during a new run, while online methods face limitations in intervention timing, knowledge guidance, and verifier availability.
    By moving offline optimization one level up, \method learns a reusable Meta-Skill from contrastive training rollouts and freezes it to guide task-local harness evolution at checkpoints online, without access to verifier feedback at test-time. 
    Across both evaluated benchmarks, this design yields the highest performance among the compared methods.
    }
  \label{fig:teaser-figure}
  \vspace{0pt}
\end{figure}


This question is difficult because evolution during a run is a sequential decision under partial information. 
At a checkpoint, an evolver can observe the active trajectory and permitted workspace state, but not the run's eventual verifier outcome. 
It must decide whether intervention is warranted, what harness component to change, and how to express the change.
Acting can rescue a failing trajectory, but it can also interrupt productive work or introduce a new failure; sometimes the correct action is to do nothing. 
Offline supervision for learning such guidance is also coarse: a terminal pass or failure evaluates a sequence of checkpoint decisions jointly rather than identifying which individual edit helped or harmed.
Finally, reasoning about every evolution decision from scratch adds online token, latency, and monetary cost.

We introduce \method, a precompute-and-reuse approach which moves offline optimization one level up. 
Rather than optimizing a shared harness, it precomputes \emph{reusable evolution knowledge} to guide the \emph{task-local edits} online.
Offline, \method learns a textual Meta-Skill $\mathcal{M}$ from training trajectories and verifier outcomes. 
Online, each run starts from the same seed harness $h^{(0)}$, and $\mathcal{M}$ stays fixed across runs.
A task agent executes normally; at checkpoints, a separate evolver uses $\mathcal{M}$ and active-run evidence $\mathcal{I}^{(k)}$ to choose $\textsc{Keep}$ or $\textsc{Evolve}(\delta^{(k)})$. 
An edit $\delta^{(k)}$ is applied through a notice, instruction update, hook, or tool channel, and affects only the remainder of the current run. 
The task agent then continues with the same environment state and partial trajectory. 
Thus, the Meta-Skill is reused across tasks, while harness edits remain local to the run and are discarded when each run ends.
Figure~\ref{fig:teaser-figure} illustrates this process.

The Meta-Skill is learned from comparisons of rollout outcomes rather than successful traces alone.
For each training task, we run the same task agent under three conditions: A, without an evolver; B, with an evolver guided by the current version of $\mathcal{M}$; and C, with an evolver given privileged verifier information available only during training.
A--B comparisons show when runs with the current evolver succeed or fail relative to runs without it; B--C comparisons reveal both recoveries missed by the current evolver and cases where verifier-informed evolution is associated with worse outcomes.
These contrasts provide directional evidence to \texttt{act}, \texttt{avoid}, or \texttt{hold} without establishing the causal effect of any single checkpoint decision.
Across training batches, \textsc{Reflect} extracts lessons from these comparisons, \textsc{Aggregate} merges recurring patterns across tasks and removes instance-specific details, and \textsc{Update} revises $\mathcal{M}$.
During online evolution, $\mathcal{M}$ is never updated with held-out verifier outcomes or task-specific edits.

We evaluate \method on Terminal-Bench 2.1 and DeepSWE 1.1. 
With GPT-5.6-Terra, it achieves the highest performance among the evaluated methods on both benchmarks, reaching 81.00\% and 44.44\%, respectively.
It also ranks first on Terminal-Bench with DeepSeek-V4-Flash-0731, scoring 77.67\%.
Ablations show gains from adding the online evolver and further gains from contrastive Meta-Skill learning.
We also report one-time offline construction and online deployment costs separately and examine how checkpoint frequency trades responsiveness for cost. 
Our contributions are:
\begin{enumerate}[leftmargin=*]
  \item We formulate harness evolution by edit timing, available evidence and the executions affected. 
  This taxonomy distinguishes offline optimization from pre-run, within-run, and after-run online adaptation.
  \item We develop \method, a hybrid harness evolution paradigm which moves offline optimization from a shared harness to reusable evolution knowledge and uses that knowledge to guide task-local harness edits within a run.
  \item We evaluate performance, cost, component contributions, and checkpoint-frequency sensitivity on two long-horizon benchmarks, with an additional backbone evaluation on Terminal-Bench.
\end{enumerate}

\section{Background: Harness Evolution}
\label{sec:formulation}

\textbf{Task execution through a harness.} (1) A task instance $\xi=(x,\mathcal{E}_x,V_x)\sim\mathcal{D}$ comprises a public instruction $x$, an interactive environment $\mathcal{E}_x$, and a verifier $V_x$ whose criteria and outcome need not be observable to the agent or the evolver.
(2) A task agent combines a model parameterized by $\theta$ with a harness $h_t$, comprising prompts, context policy, tools, skills, memory, and hooks that shape execution by, for example, preserving constraints, enabling actions, and validating artifacts.
(3) At environment step $t$, $\mathcal{E}_x$ has state $s_t$, including the current workspace and task artifacts, and emits an observation $o_t$. 
Conditioned on the instruction, interaction history, and active harness, the agent selects an action according to $a_t \sim
\pi_\theta\!\left(\,\cdot\mid x,\tau_{\leq t},h_t\right)$, where
$\tau_{\leq t}=(o_1,a_1,\ldots,a_{t-1},o_t)$.
The environment transitions via $P_x(\cdot\mid s_t,a_t)$, and the verifier returns $y=V_x(s_T)$ at termination.

\textbf{Harness Evolution.}
To improve the performance of a task agent under an initial harness, we let an evolver $g$ revise it from evidence. 
Starting from a seed harness $h^{(0)}$, decision $k\in\{1,\ldots,K\}$ proposes and applies an edit:
\begin{equation}
  \delta^{(k)}
  \sim
  g\!\left(\,\cdot\mid h^{(k-1)},\mathcal{I}^{(k)}\right),
  \qquad
  h^{(k)}
  =
  \operatorname{Apply}\!\left(h^{(k-1)},\delta^{(k)}\right),
  \label{eq:harness-evolution}
\end{equation}
where $\mathcal{I}^{(k)}$ is the evidence available to this decision, $\mathcal{M}$ is reusable knowledge about when and how edits help or harm (possibly empty), $h^{(k)}$ is the harness after decision $k$, and a null edit sets $h^{(k)}=h^{(k-1)}$. 
We evaluate harness evolution by $V_x(s_T)$ on held-out tasks. 

\textbf{Offline evolution} applies \eqref{eq:harness-evolution} on a training set $\mathcal{S}_{\mathrm{tr}}$ before any target task is observed.
Its evidence consists of completed training trajectories and their verifier outcomes, and it outputs a single harness $h^\star$ shared by all target tasks. 
This amortizes experience across training tasks, but $h^\star$ is fixed once execution begins: it cannot react to evidence revealed during a target run, and its persistent edits may overgeneralize beyond the failures that motivated them.

\textbf{Online evolution} applies \eqref{eq:harness-evolution} after the target task becomes available. There are three deployment-time settings, distinguished by when an edit is selected and which executions it affects: \emph{pre-run} selects edits before execution ($t=0$), using the target instruction and initial environment information to adapt the harness for the upcoming run; \emph{within-run} selects edits at checkpoints during execution, using the partial trajectory and permitted workspace observations to adapt the harness for the remainder of the active run; and \emph{after-run} selects edits once an attempt terminates, using completed trajectories and outcome feedback when available to adapt the harness for subsequent attempts or tasks. Only within-run evolution can respond while the motivating attempt is unfinished, but its decisions cannot depend on that attempt's eventual verifier outcome.
Appendix Table~\ref{tab:appendix-evolution-regimes} compares the two instantiations of \eqref{eq:harness-evolution}, which differ in available evidence and edit lifetime.

\textbf{The two regimes are complementary.} Offline evolution has verified training experience but cannot observe the target run, whereas within-run online evolution observes the target run but may lack reusable guidance from verified training experiences. 
This raises the question of \emph{whether verified offline experience can guide online edits without carrying tasks-specific edits across runs}.

\FloatBarrier

\section{\method: Precomputing Reusable Knowledge for Online Evolution}
\label{sec:method}

\begin{figure}[t]
    \centering
    \includegraphics[width=\linewidth]{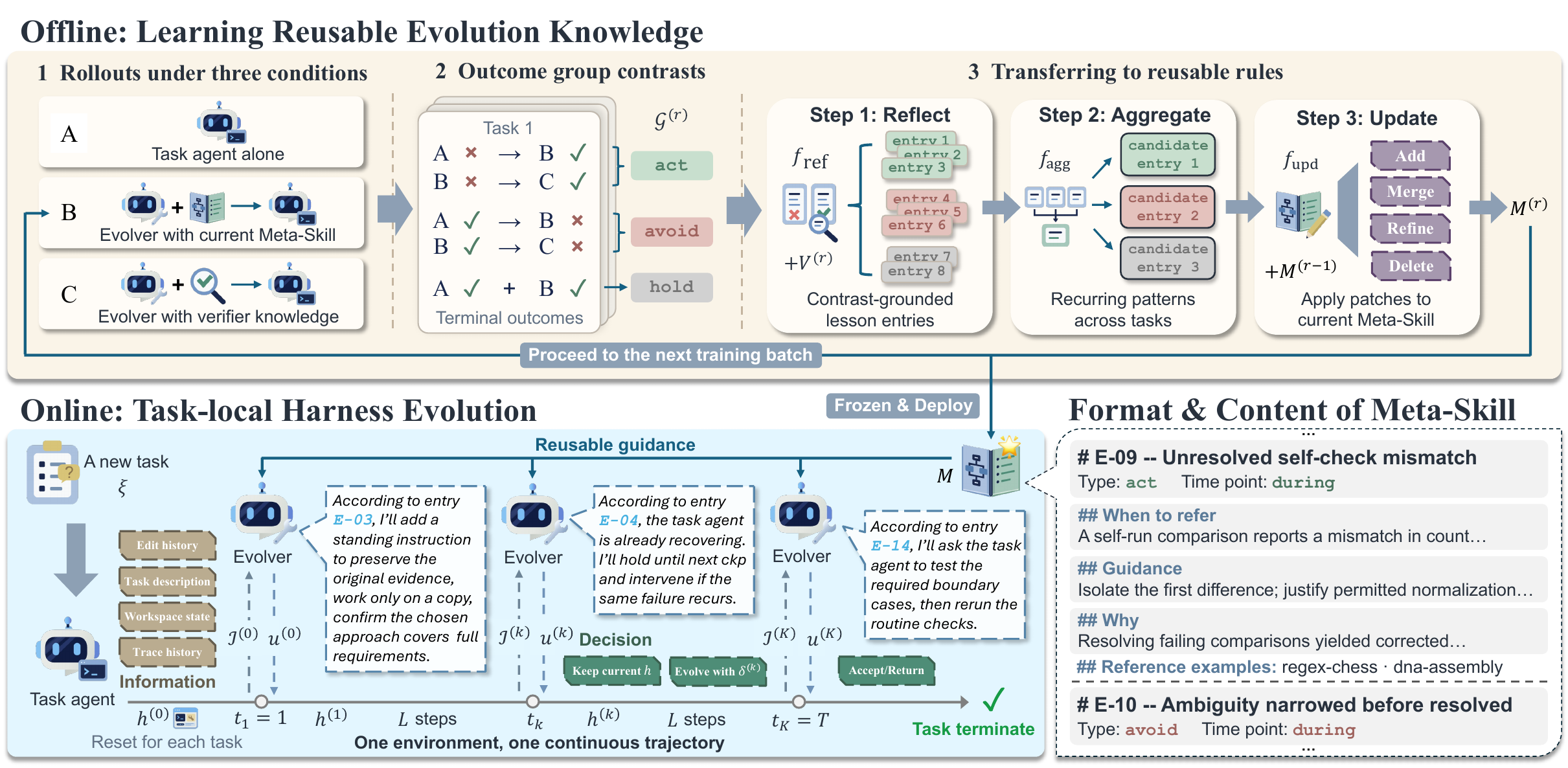}
    \caption{\textbf{Design of \method: offline knowledge learning, online harness evolution and format of Meta-Skill.}
Offline, rollouts are collected under three conditions: the task agent alone (A), evolution guided by the current Meta-Skill (B), and evolution with verifier knowledge (C).
Deterministic pairing of terminal outcomes yields \texttt{act}, \texttt{avoid}, and \texttt{hold} contrasts.
LLM-based reflection and cross-task aggregation distill these contrasts into reusable lessons, which iteratively update the Meta-Skill.
Online, the frozen Meta-Skill guides an evolver LLM to retain or modify the harness at deterministic checkpoints, while the task agent continues along a single trajectory in one environment.
Meta-Skill entries specify when, how and why to intervene  with reference examples.}
    \label{fig:experience-construction}
    \vspace{0pt}
\end{figure}

\method answers this question by moving the offline stage one level up.
Instead of optimizing a shared harness, it learns reusable evolution \emph{Meta-Skill} $\mathcal{M}$ from training trajectories and verifier feedback.
At deployment, $\mathcal{M}$ is frozen, and the evolver decides each within-run edit conditioned on it: $\delta^{(k)} \sim g(,\cdot\mid h^{(k-1)},\mathcal{I}^{(k)},\mathcal{M})$.
The Meta-Skill is reused across tasks, while each edit affects only its active run.
As shown in Figure~\ref{fig:experience-construction}, \method has three main components: a checkpoint-based interface for within-run evolution, an offline procedure that learns
$\mathcal{M}$, and a deployment protocol that reuses $\mathcal{M}$ without held-out verifier feedback.

\subsection{Online Component: Checkpoint-Based Harness Evolution}
\label{sec:evolver-task-system}

\textbf{Challenge: intervene without restarting.}
The evolver must decide whether and how to intervene before an active attempt ends, without resetting its environment, discarding its trajectory, or taking over its actions. 
\method therefore separates two roles instantiated from the
same frozen language model. 
The \textbf{task agent} starts from a minimal seed harness $h^{(0)}$ and follows $\pi_\theta$, while the \textbf{evolver} $g$ observes the run at checkpoints and may edit the harness.

\textbf{Checkpoint information.}
Let $\omega^{(k)}$ denote the permitted read-only workspace observations obtained at checkpoint $k$ which may be empty. 
The evolver also receives the newly accumulated trajectory $\Delta\tau^{(k)}$ while retaining its preceding checkpoint conversation.
Its information therefore accumulates as
\begin{equation}
  \mathcal{I}^{(k)}
  =
  \begin{cases}
    \bigl(x,\Delta\tau^{(1)},\omega^{(1)}\bigr),
    & k=1, \\
    \mathcal{I}^{(k-1)}
    \oplus
    \bigl(\Delta\tau^{(k)},\omega^{(k)},\sigma^{(k-1)}\bigr),
    & 1<k\leq K.
  \end{cases}
  \label{eq:checkpoint-information}
\end{equation}
where $\oplus$ denotes chronological append and $\sigma^{(k-1)}$ records whether the preceding edit was loaded, retained, or rolled back. 
The trajectory deltas reconstruct $\tau_{\leq t_k}$, while $\omega^{(k)}$ and $\sigma^{(k-1)}$ provide workspace observations and previous-edit status.
Appendix Table~\ref{tab:appendix-evolution-regimes} summarizes the available evidence across evolution regimes.

\textbf{Decision and continuation.}
At checkpoint $k$, the evolver makes a binary top-level decision:
\begin{equation}
  u^{(k)}
  \in
  \left\{
    \textsc{Keep},
    \textsc{Evolve}\bigl(\delta^{(k)}\bigr)
  \right\}.
  \label{eq:evolver-action-space}
\end{equation}
\textsc{Keep} sets $h^{(k)}=h^{(k-1)}$ and continues execution; 
\textsc{Evolve} applies $\delta^{(k)}$ through a bounded notice, instruction update, hook, or tool channel. 
At the terminal checkpoint, triggered by the task agent's completion claim, the evolver also receives automatically collected workspace state.
The action space remains unchanged:
\textsc{Keep} accepts the artifact, whereas \textsc{Evolve} applies a bounded correction and sends the task agent back for more work.
This follows Equation~\ref{eq:harness-evolution} as the same-run loop $\bigl(h^{(k-1)},\mathcal{I}^{(k)}\bigr)
\!\rightarrow\!u^{(k)}\!\rightarrow\!h^{(k)}$.
Appendix~\ref{app:checkpoint-interface} specifies the checkpoint schedule, observation permissions, terminal state, and edit semantics.

\subsection{Offline Component: Precomputing Reusable Evolution Knowledge}
\label{sec:meta-skill}

\textbf{Challenge: learn checkpoint guidance from terminal outcomes.}
At deployment, the evolver must choose whether and how to edit using evidence available during the run, without held-out verifier feedback.
Offline, the verifier supplies an outcome only after the entire run, evaluating a sequence of checkpoint decisions jointly. 
A passing or failing rollout therefore does not identify which intervention helped, harmed, or was unnecessary. 
Even when a useful intervention is found, its lessons must be separated from task-specific details before they can transfer.

\textbf{Optimize the knowledge, not a single harness.}
We hold the language model, seed harness, checkpoint interface, and edit channels fixed, and optimize only the evolver's reusable knowledge $\mathcal{M}$. 
This Meta-Skill links observable execution conditions to three guidance types:
\texttt{act} supports an \textsc{Evolve} decision, \texttt{hold} supports \textsc{Keep}, and \texttt{avoid} rules out a specific intervention even when another edit may remain appropriate. 
The evolver uses this guidance to decide whether and how to edit the harness during a new run.
The offline output is therefore reusable decision knowledge rather than a universal harness or a predetermined edit.
Offline optimization therefore transfers decision knowledge, not a universal harness or a predetermined edit.
Appendix~\ref{app:meta-skill-construction} gives the full offline construction algorithm, and Appendix~\ref{app:meta-skill-entries} specifies the entry fields and content of the Meta-Skill.

\textbf{A/B/C rollouts make terminal feedback informative.}
For each training task, we run the same task agent with the same language model and seed harness under three conditions: \emph{A}, without an evolver; \emph{B}, with an evolver guided by the current version of $\mathcal{M}$; and \emph{C}, with an evolver given privileged verifier information available only during training.
A--B contrasts compare outcomes with and without the current evolver, whereas B--C contrasts reveal additional recoveries or failures under verifier-informed evolution.
Condition C is particularly useful on difficult tasks, where both A and B may fail and their comparison provides no successful examples.
By giving the evolver privileged verifier information, C can produce successful rollouts that allow B--C contrasts to reveal recovery strategies missed by the current evolver.
To prevent these recoveries from relying on direct verifier leakage, the condition-C evolver's system prompt strictly prohibits passing privileged verifier information to the task agent.
We also audit condition-C trajectories after execution and exclude those that violate this restriction from contrast construction and Meta-Skill learning.

\textbf{Five contrast directions.}
B-pass/A-fail and C-pass/B-fail provide \texttt{act} evidence for potentially useful interventions and recoveries missed by the current evolver, respectively. 
B-fail/A-pass and C-fail/B-pass provide \texttt{avoid} evidence for potential harm from current and verifier-informed interventions, respectively.
A-pass/B-pass provides \texttt{hold} evidence from cases where the task agent succeeds with or without an evolver.
These comparisons give each candidate lesson a direction but do not establish the causal effect of a single checkpoint decision. 


\textbf{From task-level contrasts to reusable rules.}
We process training tasks in batches. 
For batch $r$, \textsc{Reflect} inspects each retained contrastive group and writes candidate entries following the predefined entry schema for each task; \textsc{Aggregate} merges recurring patterns across tasks and removes instance-specific details; and \textsc{Update} adds, merges, refines, or deletes entries relative to $\mathcal{M}^{(r-1)}$. 
Let $\mathcal{G}^{(r)}$ denote the retained contrastive groups and $V^{(r)}$ their privileged training-time verifier information. 
The update is
\begin{equation}
  \mathcal{M}^{(r)}
  = f_{\mathrm{upd}}\!\left(
    \mathcal{M}^{(r-1)},
    f_{\mathrm{agg}}\!\left(
      f_{\mathrm{ref}}\!\left(
        \mathcal{G}^{(r)},V^{(r)}
      \right)
    \right)
  \right).
  \label{eq:meta-skill-update}
\end{equation}
All three stages use LLM calls to the same base model used by the task agent and evolver. After the final batch, the resulting Meta-Skill is frozen for deployment, where it guides edits selected from evidence in each active run.

\subsection{Deployment Component: Frozen Knowledge and Task-Local Edits}

\textbf{Challenge: reuse without contamination.}
At deployment, the evolver receives no held-out verifier feedback to confirm whether a task-local edit helped the active run.
An edit motivated by one task's trajectory may also be inappropriate for another, so carrying it forward risks imposing task-specific changes on unrelated runs.
We therefore reuse the offline Meta-Skill but reset the harness to the same $h^{(0)}$ at the start of every new run.

\textbf{Freeze globally, adapt locally.}
At deployment, the learned Meta-Skill $\mathcal{M}$ is frozen, and every run starts from the same seed harness $h^{(0)}$.
At checkpoint $k$, the evolver decides an edit conditioned on both $\mathcal{M}$ and active-run evidence: $\delta^{(k)} \sim g(\cdot\mid h^{(k-1)},\mathcal{I}^{(k)},\mathcal{M})$. 
This extends \eqref{eq:harness-evolution} by explicitly conditioning $g$ on the reusable knowledge learned offline.
The checkpoint interface remains unchanged: \textsc{Keep} leaves the harness unchanged, while \textsc{Evolve} applies $\delta^{(k)}$ and resumes the task agent with the same environment state and partial trajectory, as described in Section~\ref{sec:evolver-task-system}.
Each edit affects only the remainder of the active run and is discarded when that run ends.
Online decisions, task-specific edits, and outcomes are not written back to $\mathcal{M}$; neither held-out verifier feedback nor an LLM-as-a-judge signal is used to update it.
The next run therefore reuses the same Meta-Skill and starts again from $h^{(0)}$, carrying over evolution knowledge but none of the preceding run's harness edits.
\section{Experiments}
\label{sec:exp}

We organize the evaluation around four questions. \textbf{RQ1} asks whether
\method improves held-out task performance. \textbf{RQ2} examines whether the
offline cost of precomputing reusable evolution knowledge is justified when
reused at deployment. \textbf{RQ3} studies the tradeoff
between checkpoint responsiveness and online cost.
\textbf{RQ4} identifies which method components account for the gains.

\subsection{Experimental Setup}

\textbf{Benchmarks and metrics.}
We evaluate \method on two challenging long-horizon benchmarks: Terminal-Bench 2.1~\citep{merrill2026terminal} and DeepSWE 1.1~\citep{huang2026deepswe}. We randomly split each benchmark into training and test sets at a 1:2 ratio. We report mean@5 over five independent runs on every held-out Terminal-Bench task and mean@3 over three independent runs on every held-out DeepSWE task.
Appendix~\ref{app:datasets} provides dataset splits and evaluation details.

\textbf{Baselines.}
We compare \method against three groups of baselines: \textbf{(1) human-designed harnesses}: NexAU~\citep{nexau2025}, Codex CLI~\citep{openai2025codex}, mini-swe-agent~\citep{yang2024swe}, Pi agent~\citep{pi_coding_agent}, and DeepSeek Harness~\citep{deepseek-harness2026}; \textbf{(2) self-evolution methods}: ACE~\citep{zhang2026agentic}, Training-free GRPO~\citep{cai2025training}, and ReasoningBank~\citep{ouyang2026reasoningbank}; and \textbf{(3) harness-evolution methods}: AHE~\citep{lin2026agentic}, Meta-Harness~\citep{lee2026meta}, and Continual Harness~\citep{karten2026continual}. All self-evolution and harness-evolution methods start from the same NexAU seed harness, denoted $h^{(0)}$ for \method.

\textbf{Implementation details.}
Self-evolution baselines construct their experience libraries on the training split; ReasoningBank and ACE can further update their libraries as test evaluation proceeds.
AHE uses best-of-two proposals for five iterations, while Meta-Harness uses one proposal per iteration for ten iterations.
Both deploy the harness with the highest training score at test-time.
Continual Harness starts from its training-evolved harness and retains edits made during test evaluation, whereas \method starts every run from $h^{(0)}$ with a frozen Meta-Skill. 
No method that adapts at test time is allowed to observe held-out verifier outcomes.
Appendix~\ref{app:baselines} gives the detailed full settings.

\textbf{LLM backbones.}
Our primary experiments use GPT-5.6-Terra~\citep{openai2026gpt56} with medium reasoning effort. 
We additionally evaluate Terminal-Bench 2.1 with DeepSeek-V4-Flash-0731~\citep{deepseekai2026deepseekv4} at high reasoning effort. 
Within each backbone setting, all agent roles across all compared systems use the same underlying model. We route all calls through one LiteLLM proxy and compute cost from the per-call charges in its response headers.

\subsection{RQ1: Does \method Improve Task Performance?}

\textbf{Primary comparison.}
With GPT-5.6-Terra, \method achieves the highest performance point estimate among evaluated methods on both benchmarks, reaching 81.00\% on Terminal-Bench 2.1 and 44.44\% on DeepSWE 1.1 as shown in Table~\ref{tab:main-results}.
These correspond to gains of 14.67 and 9.77 percentage points over the shared NexAU seed harness, compared with 7.00 and 2.66 points for Meta-Harness, the strongest harness-evolution baseline.
All evolution-based methods improve over the seed on Terminal-Bench.
On DeepSWE, however, ReasoningBank, ACE, and Continual Harness fall below the seed, while AHE only matches it. 
In contrast, \method still retains a 9.77-point improvement, suggesting that condition C can supply useful training examples on difficult tasks where other evolution methods fail to improve.

\textbf{Robustness across backbones.}
Table~\ref{tab:tb-deepseek} repeats the Terminal-Bench evaluation with DeepSeek-V4-Flash-0731 as the backbone.
\method again achieves the highest performance, reaching 77.67\% and outperforming the shared NexAU seed and Continual Harness by 7.67 and 3.00 percentage points, respectively.
The three self-evolution baselines score below the seed; AHE and Meta-Harness improve by 1.33 and 1.00 points, while Continual Harness improves by 4.67 points.
We construct a separate Meta-Skill for each backbone, so this experiment evaluates robustness across backbones rather than zero-shot transfer of a frozen Meta-Skill between models.

\begin{table}[t]
  \centering
  \footnotesize
  \caption{\textbf{Main performance and cost comparison.} Results with
    GPT-5.6-Terra (medium reasoning) on Terminal-Bench 2.1 (mean@5) and
    DeepSWE 1.1 (mean@3). Parentheses report the point change from the NexAU seed; costs are aggregate US dollars over each evaluation.}
  \label{tab:main-results}
  \vspace{0pt}
  \definecolor{tablelightblue}{rgb}{0.90,0.93,0.99}
  \definecolor{tablelightbrown}{rgb}{0.99,0.96,0.93}
  \definecolor{performanceup}{RGB}{180,75,40}
  \definecolor{performancedown}{RGB}{42,126,132}
  {%
    \newcommand{\perfgain}[1]{\kern0.06em{\scriptsize\textcolor{performanceup}{(#1)}}}
    \newcommand{\perfloss}[1]{\kern0.06em{\scriptsize\textcolor{performancedown}{(#1)}}}
    \newcommand{\perfsame}[1]{\kern0.06em{\scriptsize\textcolor{gray}{(#1)}}}
    \newif\ifhighlightgaps
    \newcommand{\tablegap}[1]{%
      \ifhighlightgaps
        {\color{tablelightbrown}%
          \rule[-\dp\csname @arstrutbox\endcsname]{#1}{%
            \dimexpr\ht\csname @arstrutbox\endcsname+\dp\csname @arstrutbox\endcsname\relax}}%
      \else
        \hspace{#1}%
      \fi}
    \renewcommand{\arraystretch}{1.08}
    \setlength{\tabcolsep}{3.5pt}
    \small
    \begin{tabularx}{\linewidth}{l@{\tablegap{7pt}}c@{\tablegap{10pt}}
      >{\centering\arraybackslash}X@{\tablegap{7pt}}
      >{\centering\arraybackslash}X@{\tablegap{7pt}}
      >{\centering\arraybackslash}X@{\tablegap{18pt}}
      c@{\tablegap{10pt}}
      >{\centering\arraybackslash}X@{\tablegap{7pt}}
      >{\centering\arraybackslash}X@{\tablegap{7pt}}
      >{\centering\arraybackslash}X}
        \toprule[1.5pt]
        \multirow{3}{*}{\textbf{Method}} &
        \multicolumn{4}{c}{\makebox[0pt][c]{\textbf{Terminal-Bench 2.1}}} &
        \multicolumn{4}{c}{\makebox[0pt][c]{\textbf{DeepSWE 1.1}}} \\[-1pt]
        \cmidrule(lr){2-5}\cmidrule(lr){6-9}
        & \multirow{2}{*}{\textbf{Perf.}} & \multicolumn{3}{c}{\textbf{Cost (\$)}} &
        \multirow{2}{*}{\textbf{Perf.}} & \multicolumn{3}{c}{\textbf{Cost (\$)}} \\[-2pt]
        \cmidrule(lr){3-5}\cmidrule(lr){7-9}
        & & \multicolumn{1}{c}{\textbf{Offline}} & \multicolumn{1}{c}{\textbf{Online}} & \multicolumn{1}{c}{\textbf{Full}} &
        & \multicolumn{1}{c}{\textbf{Offline}} & \multicolumn{1}{c}{\textbf{Online}} & \multicolumn{1}{c}{\textbf{Full}} \\[-1pt]
        \midrule[1.2pt]

        \addlinespace[1.5pt]
        \rowcolor{tablelightblue}
        \multicolumn{9}{c}{\textbf{\textit{Human-Designed Harnesses}}} \\
        \addlinespace[1.5pt]
        NexAU Seed & 66.33 & -- & 34.09 & 34.09 & 34.67 & -- & 62.67 & 62.67 \\
        Codex CLI & 68.33 & -- & 56.59 & 56.59 & 31.11 & -- & 136.53 & 136.53 \\
        mini-swe-agent & 67.33 & -- & 41.08 & 41.08 & 29.33 & -- & 68.87 & 68.87 \\
        Pi agent & 69.00 & -- & 37.98 & 37.98 & 39.11 & -- & 82.50 & 82.50 \\
        DeepSeek Harness & 68.00 & -- & 31.90 & 31.90 & 34.22 & -- & 41.37 & 41.37 \\

        \addlinespace[1.5pt]
        \rowcolor{tablelightblue}
        \multicolumn{9}{c}{\textbf{\textit{Self-Evolution Methods}}} \\
        \addlinespace[1.5pt]
        ReasoningBank & 68.00\perfgain{+1.67} & 16.04 & 55.60 & 71.64 & 33.33\perfloss{-1.34} & 55.87 & 85.64 & 141.51 \\
        ACE & 71.33\perfgain{+5.00} & 81.55 & 64.09 & 145.64 & 32.00\perfloss{-2.67} & 238.10 & 82.29 & 320.39 \\
        TF-GRPO & 74.67\perfgain{+8.34} & 92.55 & 42.62 & 135.17 & 36.44\perfgain{+1.77} & 322.50 & 86.16 & 408.66 \\

        \addlinespace[1.5pt]
        \rowcolor{tablelightblue}
        \multicolumn{9}{c}{\textbf{\textit{Harness-Evolution Methods}}} \\
        \addlinespace[1.5pt]
        AHE & 70.67\perfgain{+4.34} & 209.44 & 38.91 & 248.35 & 34.67\perfsame{0.00} & 793.48 & 69.41 & 862.89 \\
        Meta-Harness & 73.33\perfgain{+7.00} & 160.51 & 38.34 & 198.85 & 37.33\perfgain{+2.66} & 781.56 & 89.19 & 870.75 \\
        Continual Harness & 72.67\perfgain{+6.34} & 28.75 & 83.48 & 112.23 & 33.78\perfloss{-0.89} & 46.11 & 155.61 & 201.72 \\

        \addlinespace[1.5pt]
        \noalign{\global\highlightgapstrue}
        \rowcolor{tablelightbrown}
        \textbf{\method} & \textbf{81.00}\perfgain{+14.67} & 110.19 & 64.72 & 174.91 & \textbf{44.44}\perfgain{+9.77} & 238.52 & 129.34 & 367.86 \\
        \noalign{\global\highlightgapsfalse}
        \addlinespace[1.5pt]
        \bottomrule[1.5pt]
    \end{tabularx}
  }
  \vspace{-6pt}
\end{table}

\begin{table}[ht!]
  \centering
  \footnotesize
  \caption{\textbf{Robustness across backbones.} Terminal-Bench 2.1
    test-set results with DeepSeek-V4-Flash-0731 (high reasoning), reported as mean@5. 
    Parentheses report the point change from the NexAU seed; 
    costs are aggregate US dollars over each evaluation.}
  \label{tab:tb-deepseek}
  \vspace{0pt}
  \definecolor{crossmodellightblue}{rgb}{0.90,0.93,0.99}
  \definecolor{crossmodellightbrown}{rgb}{0.99,0.96,0.93}
  \definecolor{crossmodelup}{RGB}{180,75,40}
  \definecolor{crossmodeldown}{RGB}{42,126,132}
  {%
    \newcommand{\crossgain}[1]{\kern0.06em{\scriptsize\textcolor{crossmodelup}{(#1)}}}
    \newcommand{\crossloss}[1]{\kern0.06em{\scriptsize\textcolor{crossmodeldown}{(#1)}}}
    \renewcommand{\arraystretch}{1.12}
    \setlength{\tabcolsep}{9pt}
    \small
    \begin{tabular}{lc@{\hspace{14pt}}ccc}
      \toprule[1.5pt]
      \multirow{2}{*}{\textbf{Method}} & \multirow{2}{*}{\textbf{Perf.}} &
      \multicolumn{3}{c}{\textbf{Cost (\$)}} \\[-2pt]
      \cmidrule(lr){3-5}
      & & \multicolumn{1}{c}{\textbf{Offline}} & \multicolumn{1}{c}{\textbf{Online}} & \multicolumn{1}{c}{\textbf{Full}} \\[-1pt]
      \midrule[1.2pt]

      NexAU Seed & 70.00 & -- & 16.83 & 16.83 \\

      \addlinespace[2pt]
      \rowcolor{crossmodellightblue}
      \multicolumn{5}{c}{\textbf{\textit{Self-Evolution Methods}}} \\
      \addlinespace[2pt]
      ReasoningBank & 69.33\crossloss{-0.67} & 8.08 & 21.21 & 29.29 \\
      ACE & 65.00\crossloss{-5.00} & 58.30 & 21.56 & 79.86 \\
      TF-GRPO & 66.33\crossloss{-3.67} & 37.15 & 19.86 & 57.01 \\

      \addlinespace[2pt]
      \rowcolor{crossmodellightblue}
      \multicolumn{5}{c}{\textbf{\textit{Harness-Evolution Methods}}} \\
      \addlinespace[2pt]
      AHE & 71.33\crossgain{+1.33} & 123.51 & 20.28 & 143.79 \\
      Meta-Harness & 71.00\crossgain{+1.00} & 99.55 & 19.49 & 119.04 \\
      Continual Harness & 74.67\crossgain{+4.67} & 2.84 & 35.98 & 38.82 \\

      \addlinespace[2pt]
      \rowcolor{crossmodellightbrown}
      \textbf{\method} & \textbf{77.67}\crossgain{+7.67} & 81.90 & 33.51 & 115.41 \\
      \addlinespace[2pt]
      \bottomrule[1.5pt]
    \end{tabular}
  }
  \vspace{-6pt}
\end{table}

\subsection{RQ2: Is Precomputation Cost-Effective When Reused?}

\textbf{Cost comparison.}
Table~\ref{tab:main-results} separates one-time offline construction cost from online deployment cost.
Compared with the offline harness-evolution methods AHE and Meta-Harness, \method achieves higher performance at lower full cost on both benchmarks.
Its full costs are \$174.91 on Terminal-Bench and \$367.86 on DeepSWE, compared with \$248.35 and \$862.89 for AHE, and \$198.85 and \$870.75 for Meta-Harness.
With the highest observed performance among evaluated methods, \method lies on the performance--cost Pareto frontier on both benchmarks, as shown in Figure~\ref{fig:pareto-front} in Appendix~\ref{app:pareto-front}.

\textbf{Reusable precomputation.}
Compared with Continual Harness, another online harness evolution method, \method incurs higher offline construction costs: \$110.19 versus \$28.75 on Terminal-Bench 2.1 and \$238.52 versus \$46.11 on DeepSWE 1.1.
However, it achieves higher performance at lower online costs: \$64.72 versus \$83.48 and \$129.34 versus \$155.61, respectively.
The offline construction cost is a one-time investment: the resulting Meta-Skill can be reused across subsequent tasks and runs without reconstruction, so additional deployments incur only online cost.
The construction cost is therefore amortized over a larger workload as the Meta-Skill is reused.

\subsection{RQ3: How Does Checkpoint Frequency Trade Responsiveness for Cost?}

\noindent
\begin{minipage}[t]{0.51\textwidth}
  \vspace{-2pt}
  \textbf{Checkpoint-interval sensitivity.}
  Figure~\ref{fig:checkpoint-sensitivity} varies the fixed checkpoint interval $L$ on Terminal-Bench 2.1. 
  A moderate interval of $L=5$ achieves the highest performance (81.00\%) and the best observed performance--cost balance, at an
  online cost of \$64.72. 
  More frequent checkpoints increase cost without improving performance: $L=3$ and $L=1$ cost \$74.22 and \$83.41 but score
  80.33\% and 79.00\% respectively. Longer intervals reduce cost modestly, reaching \$59.10 at $L=11$ and the performance compromises as well. 
  Importantly, \method outperforms every baseline at every evaluated value of $L$, showing that its advantage is robust to the checkpoint schedule.
\end{minipage}%
\hfill
\begin{minipage}[t]{0.44\textwidth}
  \vspace{0pt}
  \centering
  \vspace{-2em}
  \includegraphics[width=\linewidth]{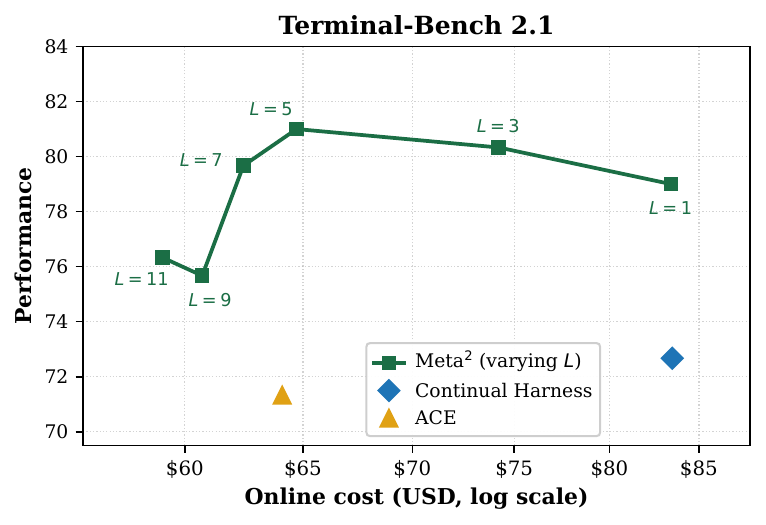}
  \vspace{-1.8em}
  \captionof{figure}{\textbf{Checkpoint-interval sensitivity.} Performance on Terminal-Bench 2.1 versus log-scaled online cost.}
  \label{fig:checkpoint-sensitivity}
\end{minipage}

\subsection{RQ4: Which Components Produce the Gains?}

\begin{figure}[ht!]
  \centering
  \includegraphics[width=0.75\linewidth]{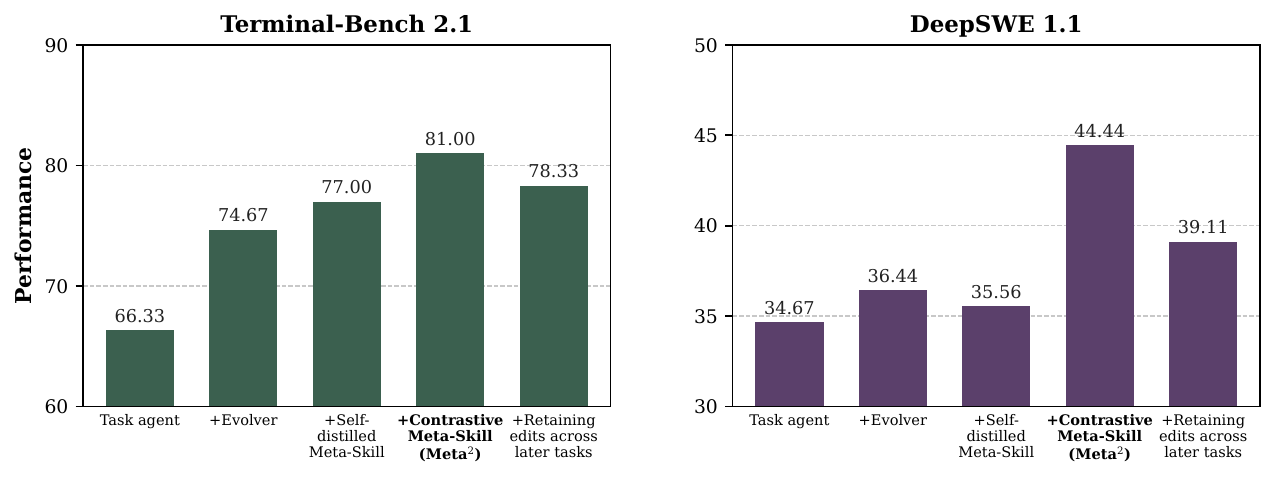}
  \setlength{\abovecaptionskip}{3pt}
  \vspace{-0.5em}
  \caption{\textbf{Component ablations.}
The first four settings evaluate the online evolver and two Meta-Skill construction methods.
The fifth retains online edits across tasks instead of resetting to $h^{(0)}$.}
  \label{fig:component-ablations}
\end{figure}

\textbf{Component contributions.}
Figure~\ref{fig:component-ablations} compares five settings.
The first four compare the task agent alone, the task agent with an online evolver without any distilled knowledge, and the two-agent system with either a self-distilled or contrastive Meta-Skill. 
Their Terminal-Bench/DeepSWE scores are 66.33/34.67, 74.67/36.44, 77.00/35.56, and 81.00/44.44, respectively.
Adding the online evolver improves performance on both benchmarks, while the self-distilled Meta-Skill improves Terminal-Bench but slightly reduces DeepSWE performance.
Contrastive Meta-Skill learning improves over self-distillation by 4.00 and 8.88 percentage points, respectively, highlighting its value particularly on harder tasks.

\textbf{Resetting between tasks.}
The fifth setting starts each new task from the harness produced by the preceding task, retaining all online edits rather than resetting to $h^{(0)}$.
Performance falls from 81.00\% to 78.33\% on Terminal-Bench and from 44.44\% to 39.11\% on DeepSWE.
Online edits are derived from the trajectory of the active task and may therefore encode task-specific interventions that do not transfer to the next task. 
These results support reusing the frozen Meta-Skill while resetting the harness for each new task, so that task-specific edits remain local to the run that produced them.


\section{Related Work}
\label{sec:related}

\textbf{Offline harness evolution.}
Using the taxonomy in Section~\ref{sec:formulation} and Appendix
Table~\ref{tab:appendix-evolution-regimes}, we group prior work by edit
selection time. Offline methods optimize the system around a frozen model on
training tasks and deploy the result as a fixed artifact. Early work
optimizes prompts and LM
pipelines~\citep{yang2024large,khattab2023dspy,yuksekgonul2024textgrad,agrawal2026gepa},
and later work searches over code-defined agent systems and
workflows~\citep{hu2025automated,zhang2025aflow}. For long-horizon agents,
recent methods evolve the harness itself: they rewrite the
scaffold~\citep{zhang2026darwin}, propose harnesses from prior code, scores,
and traces~\citep{lee2026meta}, tie edits to checkable
predictions~\citep{lin2026agentic}, or retain regression-tested
edits~\citep{zhang2026self}. Analyses suggest that many of these gains
compensate for recoverable execution defects of the underlying
model~\citep{yang2026one}. In all cases, the deployed harness $h^\star$ is
shared across held-out runs and cannot react to evidence revealed during a
run. \method also learns from completed training runs, but it
freezes reusable evolution knowledge $\mathcal{M}$ rather than a harness, so
the same $\mathcal{M}$ can support different edits in runs.

\textbf{Online harness evolution: pre-run and after-run.}
Pre-run methods adapt the harness once the target task is known but before
execution begins; for example, JIT-Agent synthesizes a task-specific harness
for each incoming task~\citep{zhang2026jit}. After-run methods adapt from
completed attempts. Reflexion~\citep{shinn2023reflexion} and Harness-R1~\citep{shao2026harness} use feedback or failure trajectories to improve later attempts on the same
task, while Dynamic Cheatsheet,
ACE, and TTHE update memory, guidance, or harness programs over a stream of
tasks~\citep{suzgun2026dynamic,zhang2026agentic,nie2026tthe}. In both
settings, the edit either precedes the evidence that execution reveals or
benefits only later attempts or tasks, and after-run selection often relies
on completed-run feedback or proxy signals. \method instead edits the run in
which the evidence appears, without observing that run's verifier outcome.

\textbf{Online harness evolution: within-run.}
The closest setting adapts the harness during an uninterrupted run.
Continual Harness refines prompts, skills, and memory inside a reset-free
episode~\citep{karten2026continual}; PILOT uses a supervisor to steer an
active worker~\citep{xiao2026pilot}; and Live-SWE-agent modifies its own
tools while solving a task~\citep{xia2025live}. These systems show that
within-run adaptation is feasible and useful, but they decide when and how to intervene without guidance. \method supplies this guidance.
Offline, A/B/C outcome contrasts produce a textual Meta-Skill that records when to act, hold, or avoid an intervention. 
Online, an evolver applies this frozen knowledge to the active run without held-out verifier feedback. 
Table~\ref{tab:related-work-coverage} and~\ref{tab:appendix-evolution-regimes} in Appendix~\ref{app:related-work-comparison} summarizes the distinction: \method is a hybrid that evolves reusable knowledge offline and task-local edits within the run.

\section{Conclusion}
\label{sec:conclusion}

\method moves offline optimization one level up, learning reusable knowledge for harness evolution instead of a fixed harness.
Offline rollout contrasts produce a Meta-Skill that is frozen at deployment; an online evolver uses it together with active-run evidence to decide harness edits that remain local to each run.
\method achieves 81.00\% on Terminal-Bench 2.1 and 44.44\% on DeepSWE 1.1 with GPT-5.6-Terra, and 77.67\% on Terminal-Bench with DeepSeek-V4-Flash-0731.
Ablations support contrastive Meta-Skill learning and resetting the harness between tasks.
These results support a simple principle: precompute \emph{how to adapt}, then adapt locally as the task unfolds.

\section{Acknowledgement}
Yuan, Liu, Han and Huang are supported by DARPA HR001124S0029-AIQ-FP-019, National Science Foundation TRAILS Institute (2229885). Private support was provided by Open Philanthropy and Apple. The Authors acknowledge the National Artificial Intelligence Research Resource (NAIRR) Pilot for contributing to this research result.

\nocite{han2025webshelldetection,han2026latex2layout,han2026zerotuning,han2026readbeforeyouthink}
\nocite{yuan2026promptguard,yu2026tsrbench}
\nocite{liu-etal-2026-ips,wang-etal-2025-reasoning-enhanced,10.1145/3711896.3737195}
\nocite{ma2026sparseautoencodersidentifyreasoning,liang2026localmdilocalfeature,liang2025revealingtreatmentnonadherencebias}
\nocite{li2026gracegroundedreasoningadapter}
\nocite{lu2025out,lu2025uni,lu2026one,lu2026worldcoder}
\nocite{jiang2026rocktokens,jiang-etal-2026-drp,jiang-ferraro-2026-beyond,jiang2026scribestructuredmidlevelsupervision,jiang2026bridgingreasoningtrajectoriesonpolicy}
\nocite{yu2026weak}

\bibliography{include/iclr2027_conference}
\bibliographystyle{plainnat}

\clearpage
\appendix
\raggedbottom

\section*{\textsc{Appendix Outline}}

This appendix provides notation, algorithmic specifications, experimental details, and supplementary evidence for \method.
It also presents case studies, additional discussion, a comparison of evolution regimes, and the prompts and Meta-Skills used by the system.
\\

\begingroup
\setlength{\parindent}{0pt}
\setlength{\parskip}{0pt}

\appendixoutlineentry{app:notation}

\appendixoutlineentry{app:algorithms}
\appendixoutlinesubentry{app:meta-skill-construction}
\appendixoutlinesubentry{app:checkpoint-interface}

\appendixoutlineentry{app:experiment-design}
\appendixoutlinesubentry{app:datasets}
\appendixoutlinesubentry{app:baselines}
\appendixoutlinesubentry{app:implementation_details}

\appendixoutlineentry{app:case-studies}

\appendixoutlineentry{app:supp-results}
\appendixoutlinesubentry{app:error-bars}
\appendixoutlinesubentry{app:pareto-front}

\appendixoutlineentry{app:discussion}
\appendixoutlinesubentry{app:limitations}
\appendixoutlinesubentry{app:impact}
\appendixoutlinesubentry{app:future-direction}

\appendixoutlineentry{app:related-work-comparison}

\appendixoutlineentry{app:prompts}
\appendixoutlinesubentry{app:online-supervisor-prompt}
\appendixoutlinesubentry{app:offline-system-prompt}
\appendixoutlinesubentry{app:reflection-prompt}
\appendixoutlinesubentry{app:aggregation-prompt}
\appendixoutlinesubentry{app:update-prompt}

\appendixoutlineentry{app:meta-skill-entries}
\appendixoutlinesubentry{app:meta-skill-entry}
\appendixoutlinesubentry{app:meta-skill-content}

\endgroup
\clearpage

\clearpage
\section{Notation}
\label{app:notation}

Table~\ref{tab:appendix-notation} summarizes the core notation used in the main text and Appendix~\ref{app:algorithms}.

\begin{table}[H]
    \centering
    \caption{Core notation used in the main text.}
    \label{tab:appendix-notation}
    \renewcommand{\arraystretch}{1.30}
    \setlength{\tabcolsep}{6pt}
    \begin{tabularx}{0.96\linewidth}{@{}>{\centering\arraybackslash}p{0.20\linewidth}
        >{\centering\arraybackslash}X@{}}
        \toprule[1.5pt]
        \textbf{Symbol} & \textbf{Meaning} \\
        \midrule
        $\xi=(x,\mathcal{E}_x,V_x)$ & Task instance: public instruction, interactive environment, and verifier. \\
        $s_t,o_t,a_t$ & State, observation, and action at step $t$. \\
        $\tau_{\leq t}$ & Partial trajectory through step $t$. \\
        $\pi_\theta$ & Frozen task-agent policy. \\
        $h^{(0)}$ & Seed harness. \\
        $h^{(k)}$ & Harness after decision $k$. \\
        $g$ & Evolver policy. \\
        $\Delta\tau^{(k)}$ & Newly accumulated trajectory at checkpoint $k$. \\
        $\omega^{(k)}$ & Permitted read-only workspace observations at checkpoint $k$. \\
        $\sigma^{(k-1)}$ & Status of the preceding edit: loaded, retained, or rolled back. \\
        $\mathcal{I}^{(k)}$ & Accumulated evolver information at checkpoint $k$. \\
        $\delta^{(k)}$ & Harness edit proposed at decision $k$. \\
        $u^{(k)}$ & \textsc{Keep} or \textsc{Evolve}$(\delta^{(k)})$ decision. \\
        $\mathcal{M}^{(r)}$ & Meta-Skill after offline batch $r$. \\
        $\mathcal{R}^{q}_{\xi}$ & Rollout records for task $\xi$ under condition $q\in\{\mathrm{A},\mathrm{B},\mathrm{C}\}$. \\
        $\mathcal{G}^{(r)}$ & Retained contrastive groups in offline batch $r$. \\
        $L$ & Checkpoint interval. \\
        \bottomrule[1.5pt]
    \end{tabularx}
\end{table}
\section{Algorithmic Details}
\label{app:algorithms}

This appendix specifies the offline Meta-Skill construction procedure and the online checkpoint evolution loop described in Section~\ref{sec:method}. 
Prompt templates are provided separately in Appendix~\ref{app:prompts}.

\subsection{Offline Meta-Skill Construction}
\label{app:meta-skill-construction}

For each training task $\xi$, let $\mathcal{R}^{q}_{\xi}$ denote the
rollout records under condition $q\in\{\mathrm{A},\mathrm{B},\mathrm{C}\}$,
including trajectories, terminal verifier outcomes, and intervention
logs when applicable. Condition A uses the task agent alone; condition B
uses an evolver guided by the current Meta-Skill; and condition C uses
an evolver with privileged training-time verifier information.
Condition-A and condition-C records are collected once and reused,
whereas condition-B records are collected with
$\mathcal{M}^{(r-1)}$ for batch $r$.

\begin{algorithm}[H]
    \caption{Offline construction of reusable evolution knowledge}
    \label{alg:offline-meta-skill}
    \small
    \begin{algorithmic}[1]
        \Require Training batches $\{\mathcal{B}_r\}_{r=1}^{R}$;
        seed harness $h^{(0)}$; fixed task agent, evolver,
        checkpoint interface, and edit channels
        \Ensure Frozen Meta-Skill $\mathcal{M}$

        \State $\mathcal{M}^{(0)}\gets\emptyset$
        \For{$r=1,\ldots,R$}
            \State $\mathcal{G}^{(r)}\gets\emptyset$
            \ForAll{$\xi\in\mathcal{B}_r$}
                \State Load fixed records
                $\mathcal{R}^{\mathrm{A}}_{\xi}$ and
                $\mathcal{R}^{\mathrm{C}}_{\xi}$
                \State Collect $\mathcal{R}^{\mathrm{B}}_{\xi}$
                using $\mathcal{M}^{(r-1)}$
                \State Exclude condition-C rollouts that fail
                the verifier-leakage audit
                \State $\mathcal{G}_{\xi}\gets
                \Call{BuildContrasts}{
                    \mathcal{R}^{\mathrm{A}}_{\xi},
                    \mathcal{R}^{\mathrm{B}}_{\xi},
                    \mathcal{R}^{\mathrm{C}}_{\xi}}$
                \State $\mathcal{G}^{(r)}\gets
                \mathcal{G}^{(r)}\cup\mathcal{G}_{\xi}$
            \EndFor

            \State $\mathcal{C}^{(r)}\gets\emptyset$
            \ForAll{retained task--direction groups
                $G\in\mathcal{G}^{(r)}$}
                \State $\mathcal{C}^{(r)}\gets
                \mathcal{C}^{(r)}\cup
                \Call{Reflect}{G,V(G)}$
            \EndFor
            \State $\widetilde{\mathcal{C}}^{(r)}\gets
            \Call{Aggregate}{\mathcal{C}^{(r)}}$
            \State $\mathcal{M}^{(r)}\gets
            \Call{Update}{
                \mathcal{M}^{(r-1)},
                \widetilde{\mathcal{C}}^{(r)}}$
        \EndFor
        \State Run training-only misfire, provenance,
        and leakage audits
        \State Freeze $\mathcal{M}\gets\mathcal{M}^{(R)}$
        \State \Return $\mathcal{M}$
    \end{algorithmic}
\end{algorithm}

\textbf{Contrast construction.}
\textsc{BuildContrasts} forms five outcome-conditioned directions:
B-pass/A-fail and C-pass/B-fail provide \texttt{act} evidence;
B-fail/A-pass and C-fail/B-pass provide \texttt{avoid} evidence;
and A-pass/B-pass provides \texttt{hold} evidence.
These contrasts provide directional evidence for candidate lessons,
without identifying the causal effect of an individual checkpoint
decision. Condition-C rollouts are excluded if privileged verifier
information is passed to the task agent.

\textbf{Reflection, aggregation, and update.}
For each retained group $G$, $V(G)$ denotes the privileged
training-time verifier information associated with its task.
\textsc{Reflect} writes candidate entries following the predefined
Meta-Skill schema. \textsc{Aggregate} groups entries by observable
trigger, removes instance-specific details, and requires support
from at least two distinct tasks. \textsc{Update} adds, merges,
refines, or deletes entries while preserving provenance.
The resulting Meta-Skill is frozen before held-out evaluation.

\subsection{Online Checkpoint Evolution}
\label{app:checkpoint-interface}

Every held-out run starts from the same seed harness $h^{(0)}$
and frozen Meta-Skill $\mathcal{M}$. Checkpoints follow the configured
schedule with interval $L$, and a completion claim triggers a
terminal checkpoint. The evolver retains its checkpoint conversation
within the active run, but receives neither held-out verifier
contents nor verifier outcomes.

\begin{algorithm}[H]
    \caption{Online evolution with a frozen Meta-Skill}
    \label{alg:online-evolution}
    \small
    \begin{algorithmic}[1]
        \Require Public task instruction $x$ and environment
        $\mathcal{E}_x$; seed harness $h^{(0)}$;
        frozen Meta-Skill $\mathcal{M}$; checkpoint interval $L$;
        task and send-back budgets
        \Ensure Final task artifact and intervention log

        \State Initialize task agent with
        $(x,\mathcal{E}_x,h^{(0)})$
        \State Initialize evolver with $\mathcal{M}$
        and an empty checkpoint conversation
        \State $k\gets0$
        \While{task budget remains}
            \State Advance the same task-agent run
            to the next checkpoint
            \State $k\gets k+1$
            \State Collect $\Delta\tau^{(k)}$
            and permitted workspace observations $\omega^{(k)}$
            \If{the task agent claims completion}
                \State Append automatically collected artifact
                and process state to $\omega^{(k)}$
            \EndIf

            \If{$k=1$}
                \State $\mathcal{I}^{(1)}\gets
                (x,\Delta\tau^{(1)},\omega^{(1)})$
            \Else
                \State Collect preceding-edit status
                $\sigma^{(k-1)}$
                \State $\mathcal{I}^{(k)}\gets
                \mathcal{I}^{(k-1)}\oplus
                (\Delta\tau^{(k)},\omega^{(k)},\sigma^{(k-1)})$
            \EndIf

            \State $u^{(k)}\gets
            \Call{Decide}{
                g,h^{(k-1)},\mathcal{I}^{(k)},\mathcal{M}}$
            \State $h^{(k)}\gets h^{(k-1)}$
            \If{$u^{(k)}=\textsc{Keep}$}
                \If{the task agent claims completion}
                    \State Accept the artifact and \textbf{break}
                \EndIf
            \Else
                \State Extract $\delta^{(k)}$ from
                $u^{(k)}=\textsc{Evolve}(\delta^{(k)})$
                \If{the task agent claims completion
                    and no send-back budget remains}
                    \State Record budget exhaustion
                    and \textbf{break}
                \EndIf
                \If{$\delta^{(k)}$ satisfies the edit contract}
                    \State $h^{(k)}\gets
                    \Call{Apply}{h^{(k-1)},\delta^{(k)}}$
                    \If{the task agent claims completion}
                        \State Consume one send-back and
                        reopen the same run for additional work
                    \EndIf
                \Else
                    \State Record edit rejection
                \EndIf
            \EndIf
            \State Continue with the same environment
            and accumulated trajectory
        \EndWhile
        \State Record the final artifact, harness versions,
        checkpoint decisions, and costs
        \State \Return Final artifact and intervention log
    \end{algorithmic}
\end{algorithm}

\textbf{Decision and edit semantics.}
\textsc{Decide} denotes the evolver's top-level
\textsc{Keep} or \textsc{Evolve}$(\delta^{(k)})$ decision,
conditioned on the current harness, accumulated checkpoint evidence,
and frozen Meta-Skill. The edit contract specifies the supporting
evidence, target, content, channel, lifetime, validation, and rollback
behavior. An edit that fails validation is rejected without changing
the harness.

\textbf{Continuation and isolation.}
Accepted edits are applied through a bounded notice, instruction
update, hook, or tool channel. A terminal \textsc{Keep} accepts the
artifact; a valid terminal \textsc{Evolve} applies a correction
and returns the task agent to the same run, subject to the send-back
budget. Neither ordinary checkpoints nor terminal send-backs reset
the environment or discard accumulated progress.
Harness edits and checkpoint conversations remain local to the run
and are discarded afterward. The Meta-Skill is never updated online,
and the held-out verifier is used only for evaluation after termination.
\section{Detailed Experimental Setup}
\label{app:experiment-design}

\subsection{Datasets and Splits}
\label{app:datasets}

We evaluate on Terminal-Bench 2.1~\citep{merrill2026terminal} and DeepSWE 1.1~\citep{huang2026deepswe}, which assess terminal-based task execution and repository-level software engineering, respectively.
Their main characteristics are summarized below.
\begin{itemize}
    \item \textbf{Terminal-Bench 2.1.}
    This benchmark contains 89 tasks requiring agents to complete complex workflows in containerized command-line environments.
    Tasks span software engineering, system administration, data processing, model training, security, and scientific computing.
    Each task provides an instruction, an execution environment, and automated verification of the resulting artifacts or system state.
    \item \textbf{DeepSWE 1.1.}
    This benchmark contains 113 original software-engineering tasks across 91 open-source repositories in TypeScript, Go, Python, JavaScript, and Rust.
    Tasks require agents to understand existing codebases and implement changes such as new features, bug fixes, and extensions to existing interfaces.
    Hand-written verifiers assess software behavior rather than requiring a particular implementation.
\end{itemize}
For each benchmark, we randomly partition the tasks into training and test sets at an approximately 1:2 ratio and fix these splits across all methods and backbone settings.
Table~\ref{tab:dataset-splits} reports the number of tasks in each split.
Offline construction of experience libraries, evolved harnesses, and Meta-Skills uses only the corresponding training split.
Held-out verifier contents and outcomes are unavailable to methods during test-time execution.

\textbf{Evaluation protocol.}
We perform five independent runs per held-out Terminal-Bench task and three per held-out DeepSWE task, reporting mean@5 and mean@3, respectively.
For each task, we average its binary verifier outcomes across repeated runs and then average these task-level scores over the test set.
This yields 300 evaluated runs per method on Terminal-Bench and 225 on DeepSWE within each backbone setting.
All methods within a benchmark/backbone setting use the same held-out tasks and repetition count.

\textbf{Execution environment.}
Task execution takes place in cloud-based sandboxes provisioned through E2B.
These sandboxes host the task environments and workspaces in which agents execute commands, inspect files, and produce task artifacts.
For \method, checkpoint interventions and terminal send-backs continue within the same sandbox, preserving the active environment state and accumulated task progress.

\begin{table}[H]
  \centering
  \caption{Task counts and evaluation repetitions for each benchmark.}
  \vspace{1pt}
  \label{tab:dataset-splits}
  \small
  \renewcommand{\arraystretch}{1.18}
  \setlength{\tabcolsep}{10pt}
  \begin{tabular}{@{\hspace{8pt}}lccc@{\hspace{8pt}}}
    \toprule[1.5pt]
    \textbf{Benchmark} & \textbf{Training tasks} & \textbf{Test tasks} & \textbf{Runs per test task} \\
    \midrule
    Terminal-Bench 2.1 & 29 & 60 & 5 \\
    DeepSWE 1.1 & 38 & 75 & 3 \\
    \bottomrule[1.5pt]
  \end{tabular}
\end{table}

\subsection{Baselines and Setup}
\label{app:baselines}
We compare against three groups of baselines: human-designed harnesses (NexAU~\cite{nexau2025}, Codex CLI~\cite{openai2025codex}, mini-swe-agent~\cite{yang2024swe}, Pi agent~\cite{pi_coding_agent}, and DeepSeek Harness~\cite{deepseek-harness2026}), self-evolution methods (ReasoningBank~\cite{ouyang2026reasoningbank}, ACE~\cite{zhang2026agentic}, and TF-GRPO~\cite{cai2025training}), and harness-evolution methods (AHE~\cite{lin2026agentic}, Meta-Harness~\cite{lee2026meta}, and Continual Harness~\cite{karten2026continual}).
All self-evolution and harness-evolution methods start from the same NexAU seed harness.

\textbf{Training-time construction.}
ACE, TF-GRPO, AHE, and Meta-Harness use five rollouts per training task for experience collection or candidate evaluation.
AHE uses best-of-two candidate proposals for five iterations on each benchmark.
Meta-Harness proposes one candidate per iteration for ten iterations.
Both AHE and Meta-Harness evaluate each candidate with five trials per training task and deploy the harness with the highest training score.
ReasoningBank and Continual Harness do not originally include a separate offline training stage.
To give them access to the same training split, we add a pre-deployment stage in which ReasoningBank accumulates experience and Continual Harness refines its harness on training tasks.
Both methods use one rollout per training task and process training tasks in windows of eight.
Their resulting experience library or evolved harness initializes test-time evaluation.

\textbf{Test-time adaptation and persistence.}
ACE and ReasoningBank continue updating their experience libraries as test evaluation proceeds, retaining these updates across runs.
Continual Harness likewise continues refining its harness and retains its edits across runs, including edits accumulated during the training stage.
Its refinement schedule uses a 25-step warmup, followed by refinement every 25 steps through step 200 and every 100 steps thereafter, with each refinement conditioned on the most recent 50 steps.
TF-GRPO, AHE, and Meta-Harness freeze their learned artifacts after training and perform no further evolution during test evaluation.
In contrast, \method reuses a frozen Meta-Skill but resets the task-agent harness to $h^{(0)}$ at the start of every run.
Its online harness edits affect only the active run and are discarded afterward.
No method that adapts at test time can access held-out verifier contents or outcomes.

\subsection{Implementation Details of \method}
\label{app:implementation_details}

\textbf{Offline Meta-Skill construction.}
For each benchmark/backbone setting, we construct a separate Meta-Skill using only the corresponding training split and freeze it before evaluation.
The task agent and evolver use the same frozen language model, and offline optimization changes only the Meta-Skill while holding the seed harness, checkpoint interface, and edit channels fixed.
We collect rollouts under three conditions: task-agent-only execution (A), evolution guided by the current Meta-Skill (B), and evolution with privileged training-time verifier information (C).
The condition-C evolver is instructed not to pass privileged verifier information to the task agent, and rollouts that fail the subsequent leakage audit are excluded from contrast construction.
Outcome-conditioned A--B and B--C comparisons provide directional evidence for candidate \texttt{act}, \texttt{hold}, and \texttt{avoid} entries.
These entries are reflected from task-level contrasts, aggregated across tasks, and used to update the Meta-Skill through the procedure in Appendix~\ref{app:meta-skill-construction}.

\textbf{Online execution.}
Every test run starts from the same NexAU seed harness $h^{(0)}$ and the frozen Meta-Skill constructed for its benchmark/backbone setting.
The evolver is first called immediately after the task agent's first step, providing an early opportunity to guide execution and avoid unnecessary subsequent exploration.
Regular checkpoints use a five-step interval, and a terminal checkpoint is triggered when the task agent stops calling tools and prepares to submit its result.
At each checkpoint, the evolver receives the newly accumulated trajectory delta, permitted read-only workspace observations, and, after the first checkpoint, the preceding-edit status.
It retains its checkpoint conversation within the active run and chooses \textsc{Keep} or \textsc{Evolve} conditioned on this accumulated evidence and the frozen Meta-Skill.
Available edit channels are a one-turn notice, a persistent instruction update, a read-only hook, and an added tool.
At the terminal checkpoint, the evolver additionally receives mechanically collected state for task-referenced paths, created or modified files, newly listening TCP ports, and processes started during the run that remain alive.
A terminal \textsc{Keep} accepts the artifact, whereas \textsc{Evolve} applies a bounded correction and sends the task agent back for additional work, subject to a bounded send-back budget.
All interventions continue within the same task-agent run and sandbox, preserving the environment state and accumulated trajectory.
Harness edits and checkpoint conversations are discarded when the run ends.
Test-time trajectories, edits, and outcomes never update the Meta-Skill, and held-out verifier contents and outcomes remain unavailable to the evolver.
\section{Case Study}
\label{app:case-studies}
\begingroup
\setkeys{Gin}{draft=false}

Each figure compares two runs of the same task: one with an evolver operating without a Meta-Skill and one with an evolver guided by the learned Meta-Skill.
Both runs use the same initial task-agent harness, language model (GPT-5.6 Terra), and checkpoint schedule, with an initial checkpoint after the first agent step, regular checkpoints every five agent steps, and a terminal checkpoint before submission.
Each lane spans the observed number of task-agent steps, and checkpoint markers indicate \textsc{Keep} or \textsc{Evolve}.
The latter includes notices, persistent instruction or hook updates, tool additions, and terminal send-backs.
Figures~\ref{fig:case-act-start}--\ref{fig:case-hold} illustrate the three Meta-Skill entry types: \texttt{act}, where guidance supports an early or mid-run intervention; \texttt{avoid}, where guidance discourages a particular intervention; and \texttt{hold}, where guidance supports leaving the harness unchanged.
Entry identifiers are specific to each benchmark's Meta-Skill.

\makeatletter
\setlength{\@fptop}{0pt}
\setlength{\@dblfptop}{0pt}
\makeatother

\captionsetup[figure]{skip=5pt,justification=justified,singlelinecheck=false}

\clearpage
\begin{figure*}[p]
  \centering
  \includegraphics[width=0.95\textwidth,height=0.38\textheight,keepaspectratio]{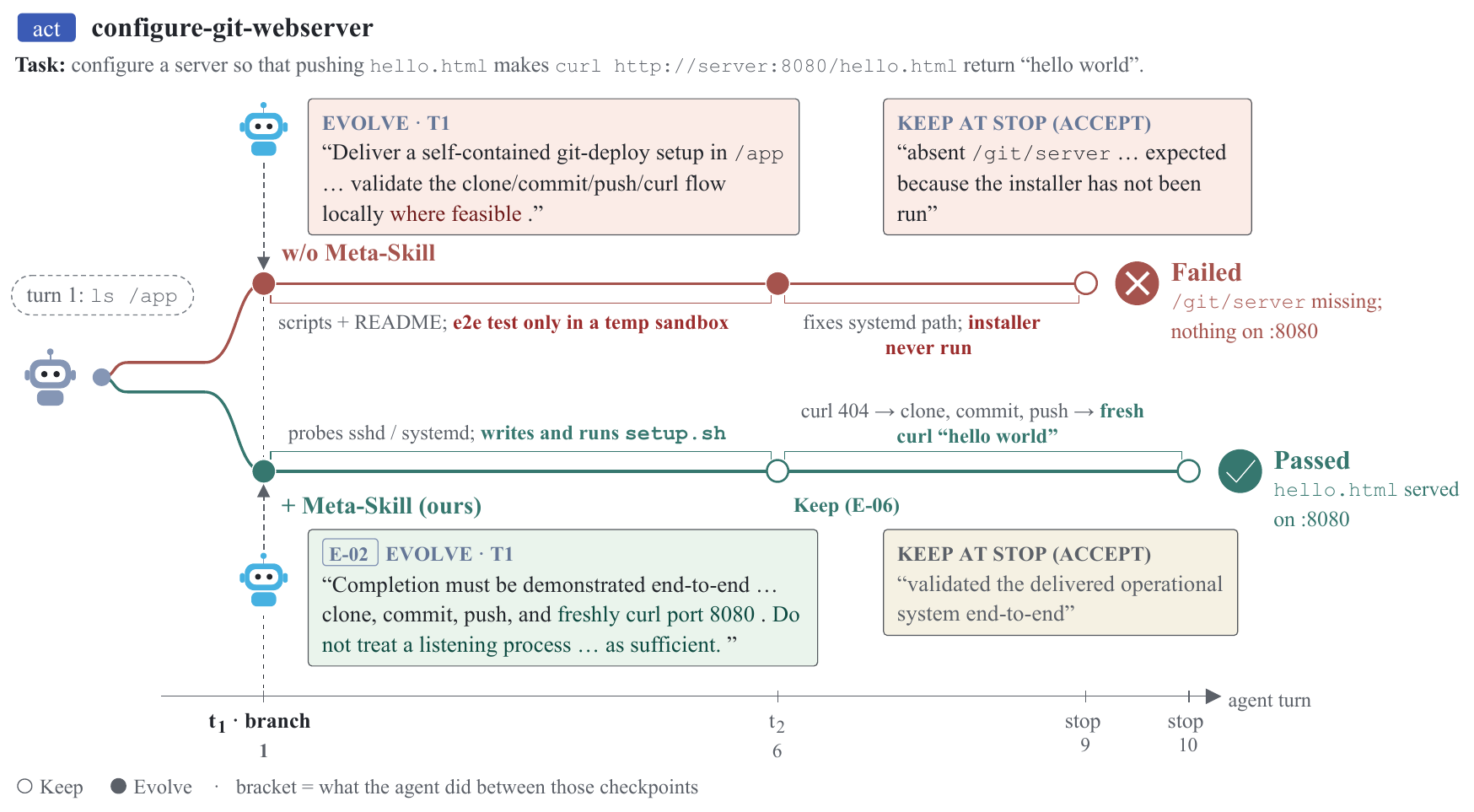}
\caption{\textbf{Early intervention guided by an \texttt{act} entry} (Terminal-Bench~2.1).
  The task requires a Git push to deploy \texttt{hello.html} to a web server on port 8080.
  At the first checkpoint, the evolver without Meta-Skill specifies an implementation and requests validation only ``where feasible.''
  The task agent tests in a temporary sandbox but does not run the installer in the task environment; the evolver accepts the result despite the missing \texttt{/git/server} directory and inactive service.
  With Meta-Skill, entry E-02 guides the evolver to require end-to-end evidence: a clone, commit, and push followed by a fresh HTTP request confirming that the deployed page returns ``hello world.''
  The task agent installs the service and verifies this sequence, after which the evolver chooses \textsc{Keep} at the subsequent and terminal checkpoints.
  }
  \label{fig:case-act-start}

  \par\vspace{2pt}

  \centering
  \includegraphics[width=0.95\textwidth,height=0.38\textheight,keepaspectratio]{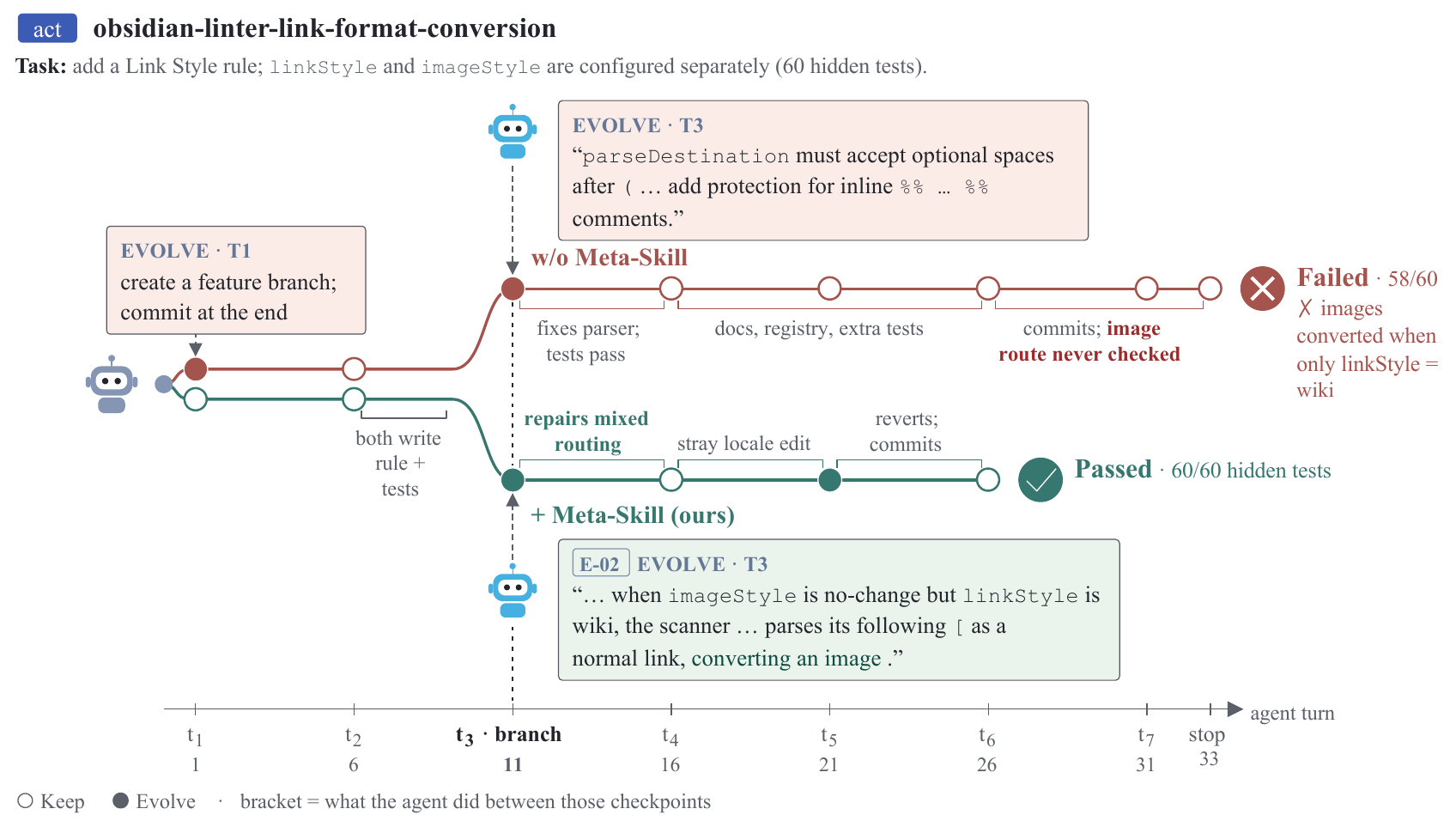}
\caption{\textbf{Mid-run intervention guided by an \texttt{act} entry} (DeepSWE~1.1).
  The task requires a Link Style rule with independently configured \texttt{linkStyle} and \texttt{imageStyle}.
  By checkpoint $t_3$, both task agents have produced an initial implementation, and both evolvers issue a notice.
  Without Meta-Skill, the notice focuses on parser issues exposed by failing tests, while the image-handling route remains unchecked.
  The submitted implementation passes 58 of 60 hidden tests, incorrectly converting images when \texttt{linkStyle} is set to \texttt{wiki} and \texttt{imageStyle} is set to leave images unchanged.
  With Meta-Skill, entry E-02 directs attention to an unverified execution route, and the evolver identifies the interaction between link and image parsing.
  The task agent repairs this route, and the submitted implementation passes all 60 hidden tests.
  }
  \label{fig:case-act}

\end{figure*}

\begin{figure*}[p]
  \centering
  \includegraphics[width=0.95\textwidth,height=0.38\textheight,keepaspectratio]{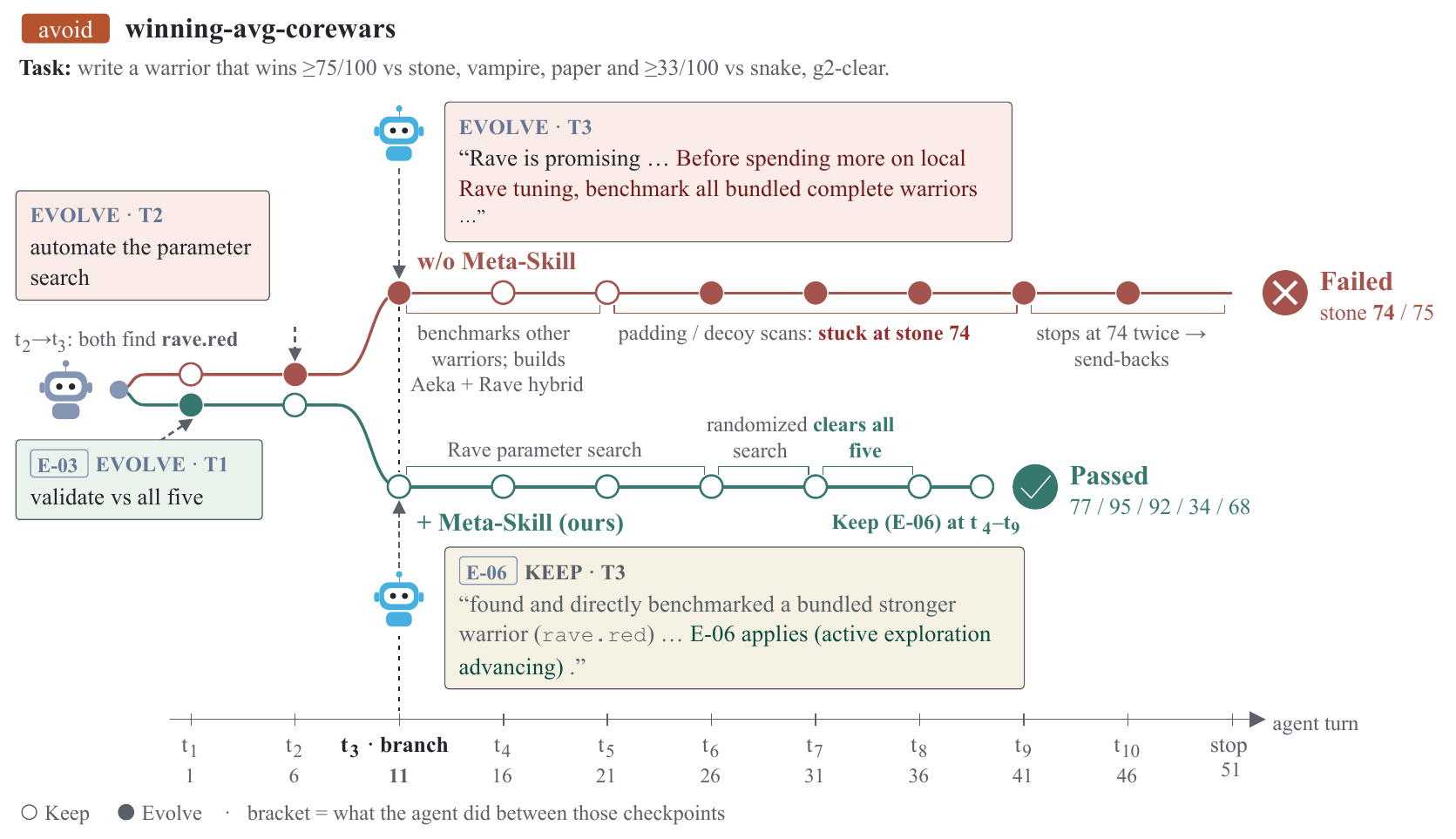}
\caption{\textbf{Avoiding premature redirection with an \texttt{avoid} entry} (Terminal-Bench~2.1).
  By checkpoint $t_3$, both task agents have identified the bundled Rave warrior as a promising candidate, but its performance against stone remains below the required threshold.
  Without Meta-Skill, the evolver uses its final notice to redirect exploration toward other bundled warriors.
  The task agent constructs an Aeka--Rave hybrid but finishes with 74 wins against stone, one short of the required 75.
  With Meta-Skill, entry E-06 discourages changing the search strategy while exploration continues to make progress.
  The evolver chooses \textsc{Keep} at $t_3$ and subsequent checkpoints, allowing the task agent to continue tuning Rave and satisfy all five performance thresholds.
  }
  \label{fig:case-avoid}

  \par\vspace{2pt}
  \includegraphics[width=0.95\textwidth,height=0.38\textheight,keepaspectratio]{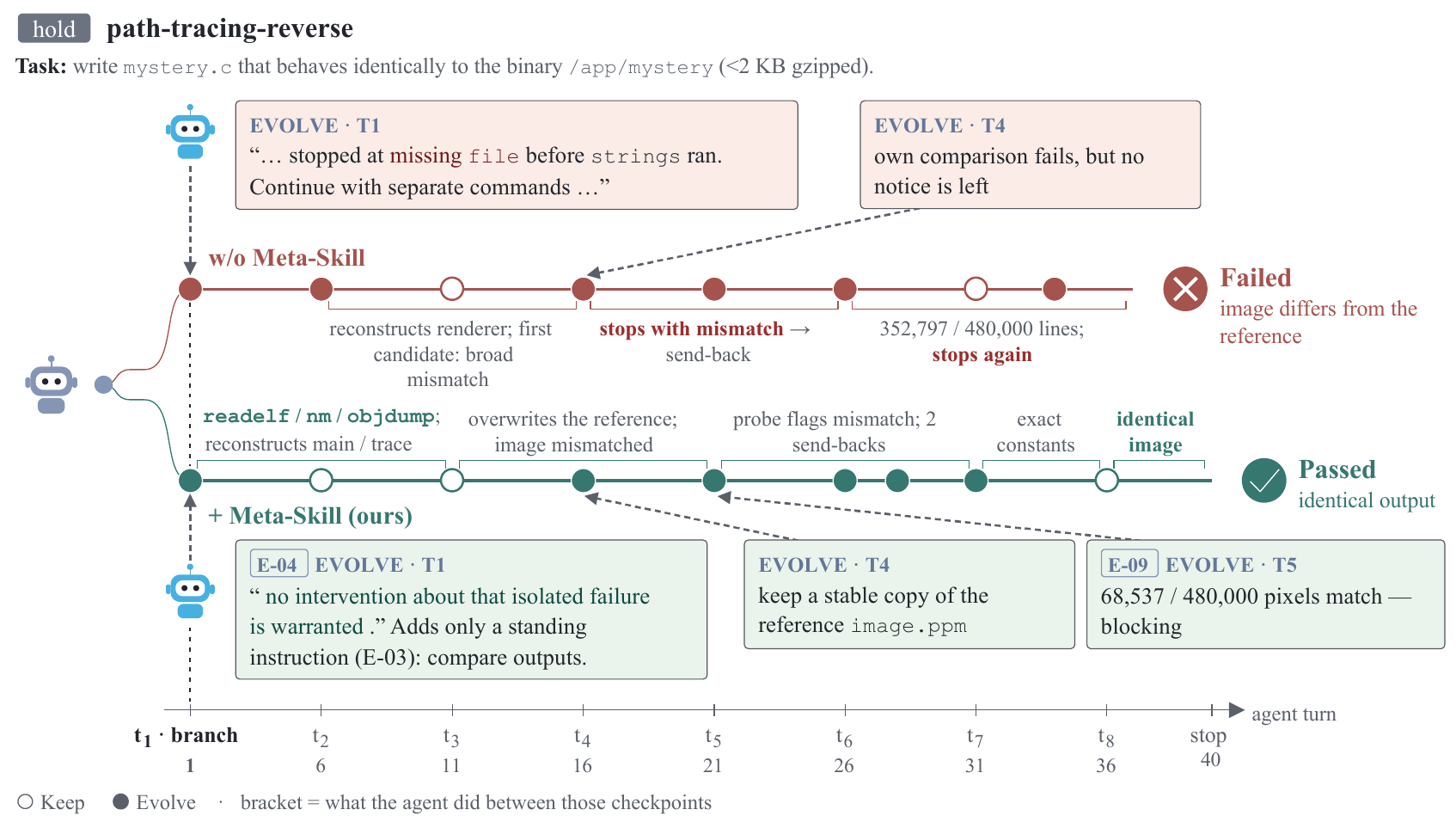}
\caption{\textbf{Preserving the notice budget with a \texttt{hold} entry} (Terminal-Bench~2.1).
  In both runs, the first command fails because the \texttt{file} utility is unavailable, and the task agent subsequently switches to alternative tools.
  Without Meta-Skill, the evolver spends notices at $t_1$ and $t_2$, exhausting this channel before output comparisons reveal a substantial mismatch.
  With Meta-Skill, entry E-04 discourages a notice about the isolated tool failure, while a separate instruction guided by E-03 requires output comparison.
  The two notices remain available at $t_4$--$t_5$ to address reference preservation and the observed mismatch, and the task agent ultimately reproduces the reference output.
  This case illustrates a difference in intervention timing rather than establishing that withholding the early notice determines success.}
  \label{fig:case-hold}
\end{figure*}

\clearpage
\endgroup

\section{Supplementary Experiment Results}
\label{app:supp-results}
\subsection{Mean and Standard Deviation Across Repeated Runs}
\label{app:error-bars}
Table~\ref{tab:reported-error-bars} reports mean performance and sample standard deviation across repeated evaluation runs.
Each repetition covers the full held-out task set: 60 tasks for Terminal-Bench 2.1 and 75 tasks for DeepSWE 1.1.
We compute the percentage of tasks passed in each repetition, then report the mean and sample standard deviation over five repetitions for Terminal-Bench and three for DeepSWE.
Standard deviations are expressed in percentage points and describe variation across repetitions.

\begin{table}[H]
    \centering
    \caption{Mean performance $\pm$ sample standard deviation across repeated evaluation runs (percentage points).
    Terminal-Bench 2.1 uses five repetitions over 60 tasks, and DeepSWE 1.1 uses three repetitions over 75 tasks.
    Terra denotes GPT-5.6-Terra, and DeepSeek denotes DeepSeek-V4-Flash-0731.
    Dashes indicate settings not evaluated.}
    \label{tab:reported-error-bars}
    \small
    \renewcommand{\arraystretch}{1.18}
    \setlength{\tabcolsep}{8pt}
    \begin{tabular}{@{\hspace{6pt}}lccc@{\hspace{6pt}}}
        \toprule[1.5pt]
        & \multicolumn{2}{c}{\textbf{Terra}}
        & \textbf{DeepSeek} \\
        \cmidrule(lr){2-3}
        \cmidrule(lr){4-4}
        \textbf{Method}
        & \textbf{Terminal-Bench 2.1}
        & \textbf{DeepSWE 1.1}
        & \textbf{Terminal-Bench 2.1} \\
        \midrule[1.2pt]
        NexAU Seed
        & $66.33 \pm 2.17$
        & $34.67 \pm 1.33$
        & $70.00 \pm 2.64$ \\
        Codex CLI
        & $68.33 \pm 5.14$
        & $31.11 \pm 3.08$
        & -- \\
        mini-swe-agent
        & $67.33 \pm 4.18$
        & $29.33 \pm 3.53$
        & -- \\
        Pi agent
        & $69.00 \pm 4.94$
        & $39.11 \pm 6.71$
        & -- \\
        DeepSeek Harness
        & $68.00 \pm 3.42$
        & $34.22 \pm 4.68$
        & -- \\
        \midrule
        ReasoningBank
        & $68.00 \pm 3.61$
        & $33.33 \pm 2.31$
        & $69.33 \pm 3.03$ \\
        ACE
        & $71.33 \pm 2.74$
        & $32.00 \pm 4.62$
        & $65.00 \pm 2.64$ \\
        TF-GRPO
        & $74.67 \pm 4.62$
        & $36.44 \pm 5.39$
        & $66.33 \pm 4.31$ \\
        \midrule
        AHE
        & $70.67 \pm 4.65$
        & $34.67 \pm 4.62$
        & $71.33 \pm 3.98$ \\
        Meta-Harness
        & $73.33 \pm 4.41$
        & $37.33 \pm 2.67$
        & $71.00 \pm 2.53$ \\
        Continual Harness
        & $72.67 \pm 4.65$
        & $33.78 \pm 5.05$
        & $74.67 \pm 4.62$ \\
        \method
        & $81.00 \pm 0.91$
        & $44.44 \pm 0.77$
        & $77.67 \pm 3.84$ \\
        \bottomrule[1.5pt]
    \end{tabular}
\end{table}

\subsection{Cost--Performance Pareto Frontier}
\label{app:pareto-front}

\begin{figure}[t]
    \centering
    \includegraphics[width=\linewidth]{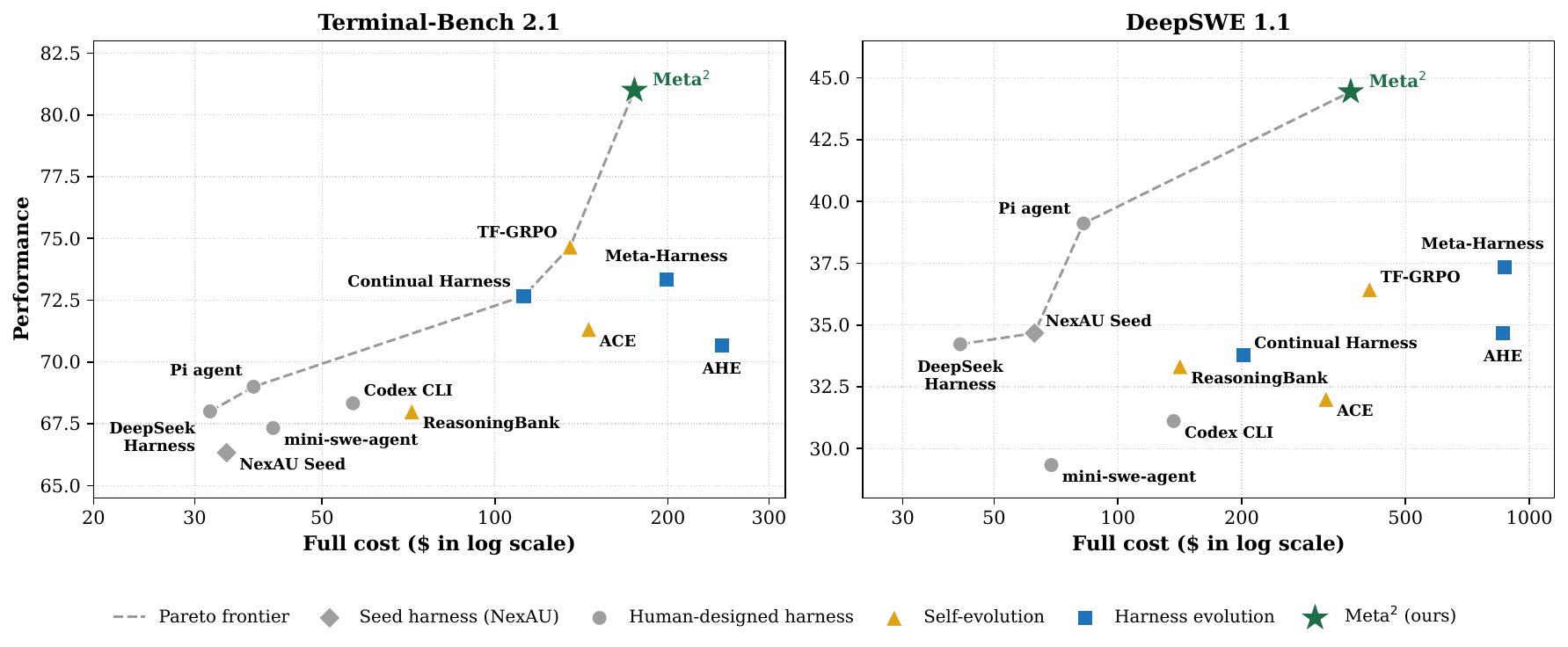}
    \caption{\textbf{Performance versus full cost with GPT-5.6-Terra.}
    Full cost includes one-time offline construction and online evaluation costs, plotted on a logarithmic horizontal scale.
    Dashed lines connect the non-dominated performance--cost point estimates for Terminal-Bench 2.1 (left) and DeepSWE 1.1 (right).}
    \label{fig:pareto-front}
\end{figure}

Figure~\ref{fig:pareto-front} compares held-out performance against full cost for GPT-5.6-Terra on both benchmarks.
Full cost includes offline construction, where applicable, and online execution over the complete held-out evaluation.
A method is Pareto dominated if another method has no higher cost and no lower performance, with a strict improvement in at least one dimension.
The dashed lines connect the non-dominated point estimates.

\method lies on the Pareto frontier and achieves the highest performance point estimate on both benchmarks.
It also dominates AHE and Meta-Harness, attaining higher performance at lower full cost.
Lower-cost human-designed harnesses remain on both frontiers, while Continual Harness and TF-GRPO provide intermediate cost--performance tradeoffs on Terminal-Bench.
Thus, \method extends the frontier toward higher performance while retaining alternatives with lower total expenditure.
These comparisons use the evaluation sizes reported in Appendix~\ref{app:datasets}; the contribution of one-time offline construction to full cost depends on the number of subsequent runs.
\section{Additional Discussion}
\label{app:discussion}

\subsection{Limitations}
\label{app:limitations}
\textbf{Interpretation of outcome contrasts.}
The A/B/C comparisons provide run-level evidence for evolution guidance rather than causal estimates of individual checkpoint decisions.
Because independently sampled trajectories may diverge before an intervention, differences in terminal outcomes cannot be attributed to a single edit.
Accordingly, the Meta-Skill records guidance grounded in observed contrasts, and the evolver assesses its applicability using evidence from the active run.
Controlled comparisons that resume from a shared checkpoint state would enable more precise attribution.

\textbf{Deployment cost and intervention timing.}
The evolver introduces additional model calls, and its deployment cost depends on checkpoint frequency and the extent of intervention.
Our checkpoint-frequency study examines this tradeoff, while \texttt{hold} and \texttt{avoid} guidance supports selective intervention.
Bounded edit channels, validation, and rollback further constrain changes to the active harness.
These mechanisms do not guarantee that every intervention is useful or timely, leaving room for more efficient checkpoint scheduling and decision policies.

\textbf{Scope of reusable guidance.}
Cross-task aggregation removes instance-specific details to produce guidance that can be reused across tasks.
The resulting entries remain abstractions of the training evidence, and their applicability depends on whether the active run exhibits the relevant conditions.
Support from multiple tasks and retained source examples aid interpretation and traceability, but do not establish applicability to all task distributions.
Evaluating transfer under broader distribution shifts is a useful direction for future work.

\textbf{Evaluation coverage.}
Our evaluation focuses on terminal and software-engineering tasks with inspectable artifacts and automatic verification.
The results establish performance within these settings and the evaluated backbone configurations.
Further evaluation is needed to assess applicability to domains with different interaction structures or less explicit completion criteria, such as embodied control and open-ended research.

\subsection{Broader Impact and Safety}
\label{app:impact}
Online harness evolution may improve the reliability of long-horizon agents by allowing them to respond to tool failures, artifact errors, and evidence revealed during execution.
A reusable Meta-Skill can support consistent intervention decisions when similar situations recur.
Because harness edits influence subsequent agent behavior, deployment should account for tool permissions, the confidentiality of trajectory data, and the traceability of changes.

Our implementation restricts workspace inspection to permitted read-only observations and applies task-local edits through bounded channels with validation and rollback.
The Meta-Skill remains frozen during deployment, and intervention logs record checkpoint decisions and harness changes.
For broader deployment, we recommend least-privilege tool access, explicit edit scopes, versioned changes, and access-controlled audit logs.
Edits that expand permissions or authorize external actions beyond the original task scope should require human approval.
Privileged verifier information is used only during offline construction, with condition-C rollouts audited for leakage before inclusion in Meta-Skill learning.

\subsection{Future Directions}
\label{app:future-direction}
One direction is to refine the Meta-Skill during deployment using reliable feedback from newly completed tasks.
This could help the guidance adapt to changing task distributions, provided that updates distinguish reusable lessons from task-specific corrections and are validated before reuse.
Such a protocol would require an explicit source of deployment feedback, separate from held-out verifier information reserved for evaluation.

A second direction is to replace fixed checkpoint intervals with an adaptive scheduling policy.
The policy could use execution evidence and evolver uncertainty to determine when inspection is warranted, balancing timely intervention against monitoring cost.
Further extensions include retrieving relevant entries from larger Meta-Skills, identifying situations not covered by existing guidance, and evaluating transfer across task-agent models and domains.

\section{Evolution Regimes and Related-Work Comparison}
\label{app:related-work-comparison}

Table~\ref{tab:related-work-coverage} summarizes the evolution regimes supported by representative methods.
The four regime columns describe offline optimization and pre-run, within-run, and after-run evolution, following the definitions in Section~\ref{sec:formulation}.
A checkmark denotes explicit support, a triangle denotes support restricted to particular modes or reuse of completed-run artifacts, and a cross denotes that the regime is not supported in the described protocol.
The final column indicates whether deployment-time adaptation requires task-verifier feedback, including feedback used to select updates for subsequent runs.
The comparison describes the original methods rather than the training-time adaptations introduced for our baseline evaluation.

\begin{table}[H]
    \centering
    \caption{\textbf{Evolution regimes and test-time verifier requirements of representative methods.}
    $\checkmark$ denotes explicit support, $\triangle$ restricted support, and $\times$ no support in the described protocol.
    Verifier requirements refer to deployment-time adaptation rather than offline training or final evaluation.}
    \label{tab:related-work-coverage}
    \small
    \renewcommand{\arraystretch}{1.20}
    \setlength{\tabcolsep}{2pt}
    \begin{tabularx}{\linewidth}{
        @{\hspace{5pt}}
        >{\raggedright\arraybackslash}p{0.22\linewidth}
        *{4}{>{\centering\arraybackslash}X}
        >{\centering\arraybackslash}p{0.19\linewidth}
        @{\hspace{5pt}}
    }
        \toprule[1.5pt]
        \textbf{Method}
        & \shortstack{\textbf{Offline}\\\textbf{optimization}}
        & \shortstack{\textbf{Pre-run}\\\textbf{evolution}}
        & \shortstack{\textbf{Within-run}\\\textbf{evolution}}
        & \shortstack{\textbf{After-run}\\\textbf{evolution}}
        & \shortstack{\textbf{Test-time verifier}\\\textbf{requirement}} \\
        \midrule[1.2pt]
        Meta-Harness~\citep{lee2026meta}
        & $\checkmark$ & $\times$ & $\times$ & $\times$
        & Not required \\
        AHE~\citep{lin2026agentic}
        & $\checkmark$ & $\times$ & $\times$ & $\times$
        & Not required \\
        JIT-Agent~\citep{zhang2026jit}
        & $\checkmark$ & $\checkmark$ & $\times$ & $\triangle$
        & Mode-dependent \\
        ACE~\citep{zhang2026agentic}
        & $\checkmark$ & $\times$ & $\times$ & $\checkmark$
        & Mode-dependent \\
        Harness-R1~\citep{shao2026harness}
        & $\checkmark$ & $\times$ & $\times$ & $\checkmark$
        & Required \\
        Continual Harness~\citep{karten2026continual}
        & $\times$ & $\times$ & $\checkmark$ & $\triangle$
        & Not required \\
        PILOT~\citep{xiao2026pilot}
        & $\times$ & $\times$ & $\checkmark$ & $\checkmark$
        & Required \\
        Live-SWE-agent~\citep{xia2025live}
        & $\times$ & $\times$ & $\checkmark$ & $\times$
        & Not required \\
        \midrule
        \method (ours)
        & $\checkmark$ & $\times$ & $\checkmark$ & $\times$
        & Not required \\
        \bottomrule[1.2pt]
    \end{tabularx}
\end{table}

The online evolution regimes are not mutually exclusive: a method may adapt before or during a run and also use completed-run evidence to update artifacts for subsequent tasks.
For example, JIT-Agent generates a task-specific harness before execution and, in its streaming mode, uses the completed rollout's reward, latency, and cost to update the harness bank.
PILOT combines within-run steering and harness updates with verifier-based selection of which updates to retain across iterations.
Continual Harness refines its harness within a run and retains the resulting harness in subsequent runs.

Table~\ref{tab:appendix-evolution-regimes} summarizes harness evolution by decision timing, available evidence, and the executions affected by an update.
Offline optimization uses training experience to construct an artifact shared across target runs, whereas online evolution adapts after the target task becomes available.
The online settings differ in whether an update affects the upcoming run, the remainder of an active run, or subsequent attempts or tasks.
The final row locates \method within this taxonomy: the Meta-Skill $\mathcal{M}$ is learned offline and reused across tasks, while checkpoint information $\mathcal{I}^{(k)}$ guides task-local edits $\delta^{(k)}$ during execution.

\begin{table}[H]
    \centering
    \caption{\textbf{Evolution regimes by timing, evidence, and update scope.}
    Online edits occur pre-run, within-run, or after-run; after-run updates target either the next attempt or later tasks. 
    \method learns reusable $\mathcal{M}$ offline and applies task-local edits within-run.}
    \label{tab:appendix-evolution-regimes}
    \vspace{4pt}
\begingroup
\definecolor{regInk}{HTML}{283B50}
\definecolor{regBlue}{HTML}{376FA6}
\definecolor{regBluePale}{HTML}{EFF5FC}
\definecolor{regOrange}{HTML}{B65B27}
\definecolor{regOrangePale}{HTML}{FFF4EC}
\definecolor{regTeal}{HTML}{287F83}
\definecolor{regGrayPale}{HTML}{F3F5F7}
\definecolor{regBrownPale}{HTML}{FDF5EF}
\resizebox{.98\linewidth}{!}{%
\begin{tikzpicture}[
    x=.96cm,y=.82cm,
    font=\sffamily\fontsize{7.5}{8.9}\selectfont,
    text=regInk,
    phase/.style={draw=regInk!24,rounded corners=2pt,line width=.55pt,
                  align=center,font=\sffamily\bfseries\fontsize{7.4}{8.8}\selectfont,
                  minimum height=.52cm},
    rowlabel/.style={anchor=east,align=right,
                     font=\sffamily\bfseries\fontsize{7.4}{8.7}\selectfont},
    scope/.style={-{Stealth[length=4.5pt,width=4pt]},line width=1.05pt,line cap=round},
    guide/.style={draw=regInk!20,dashed,line width=.5pt},
    note/.style={font=\sffamily\fontsize{7}{8.2}\selectfont,text=regInk!72},
    grouplabel/.style={font=\sffamily\bfseries\fontsize{7}{8.2}\selectfont}
]
    \node[phase,fill=regGrayPale,minimum width=1.95cm]
      at (4.25,3.73) {Training $\mathcal{S}_{\mathrm{tr}}$};
    \node[phase,fill=regBluePale,minimum width=3.55cm]
      at (7.55,3.73) {Task $\xi_j$: attempt $m$};
    \node[phase,fill=regBluePale,minimum width=2.55cm]
      at (10.85,3.73) {Attempt $m{+}1$};
    \node[phase,fill=regGrayPale,minimum width=3.05cm]
      at (14.18,3.73) {Later tasks $\xi_{j+1},\ldots$};

    \fill[regTeal] (5.20,4.28) circle (.075);
    \draw[scope,draw=regTeal,line width=1.15pt] (5.20,4.28) -- (15.75,4.28)
      node[midway,above=1.5pt,text=regTeal,
           font=\sffamily\bfseries\fontsize{7.2}{8.5}\selectfont]
      {Frozen Meta-Skill $\mathcal{M}$ reused across target tasks};

    \fill[regGrayPale,rounded corners=2pt]
      (1.55,2.42) rectangle (15.82,3.00);
    \fill[regBluePale!72,rounded corners=2pt]
      (1.55,-.20) rectangle (15.82,2.34);
    \fill[regOrangePale,rounded corners=2pt]
      (1.72,1.15) rectangle (15.67,1.69);

    \foreach \x in {5.20,9.30,12.25} {
      \draw[guide] (\x,-.20) -- (\x,3.37);
    }
    \draw[guide] (7.38,1.12) -- (7.38,3.37);
    \node[note,fill=white,inner sep=1pt] at (5.20,3.34) {$t=0$};
    \node[note,fill=white,inner sep=1pt] at (7.38,3.34) {$t_k$};
    \node[note,fill=white,inner sep=1pt] at (9.30,3.34) {$T_m$};

    \draw[regInk!35,line width=1pt] (1.34,2.48) -- (1.34,2.94);
    \draw[regInk!35,line width=1pt] (1.34,2.48) -- (1.55,2.48);
    \draw[regInk!35,line width=1pt] (1.34,2.94) -- (1.55,2.94);
    \node[grouplabel,anchor=east,text=regInk!82] at (1.16,2.71) {OFFLINE};

    \draw[regBlue!65,line width=1pt] (1.34,-.12) -- (1.34,2.28);
    \draw[regBlue!65,line width=1pt] (1.34,-.12) -- (1.55,-.12);
    \draw[regBlue!65,line width=1pt] (1.34,2.28) -- (1.55,2.28);
    \node[grouplabel,anchor=east,align=right,text=regBlue] at (1.16,1.17)
      {ONLINE\\[-1pt]EVOLUTION};

    \node[rowlabel] at (3.90,2.71) {Pre-deployment};
    \fill[regInk!65] (5.02,2.71) circle (.085);
    \draw[scope,draw=regInk!65] (5.02,2.71) -- (15.70,2.71);

    \node[rowlabel] at (3.90,2.06) {Pre-run};
    \fill[regInk!65] (5.20,2.06) circle (.085);
    \draw[scope,draw=regInk!65] (5.20,2.06) -- (9.15,2.06);

    \node[rowlabel,text=regOrange] at (3.90,1.42) {Within-run};
    \fill[regOrange] (7.38,1.42) circle (.09);
    \draw[scope,draw=regOrange,line width=1.15pt] (7.38,1.42) -- (9.15,1.42);

    \node[rowlabel] at (3.90,.77)
      {After-run\\[-1.5pt]\textcolor{regInk!62}{\mdseries\fontsize{6.6}{7.5}\selectfont same task}};
    \fill[regInk!65] (9.30,.77) circle (.085);
    \draw[scope,draw=regInk!65] (9.30,.77) -- (12.10,.77);

    \node[rowlabel] at (3.90,.13)
      {After-run\\[-1.5pt]\textcolor{regInk!62}{\mdseries\fontsize{6.6}{7.5}\selectfont task stream}};
    \fill[regInk!65] (12.25,.13) circle (.085);
    \draw[scope,draw=regInk!65] (12.25,.13) -- (15.70,.13);

    \fill[regInk!60] (4.18,-.43) circle (.075);
    \node[anchor=west,note] at (4.33,-.43) {edit selected};
    \draw[scope,draw=regInk!55] (6.35,-.43) -- (7.35,-.43);
    \node[anchor=west,note] at (7.48,-.43) {affected executions};
\end{tikzpicture}%
}
\endgroup\par
    \vspace{3pt}
    \definecolor{regimeLightBrown}{rgb}{0.99,0.96,0.93}
    \definecolor{regimeLightBlue}{rgb}{0.90,0.93,0.99}
    \definecolor{regimeLightOrange}{HTML}{FFF4EC}
    \definecolor{regimeTeal}{HTML}{287F83}
    \definecolor{regimeOrange}{HTML}{B65B27}
    {%
    \small
    \renewcommand{\tabularxcolumn}[1]{p{#1}}
    \renewcommand{\arraystretch}{1.10}
    \setlength{\tabcolsep}{4pt}
    \begin{tabularx}{\linewidth}{@{}>{\raggedright\arraybackslash}p{0.185\linewidth}
        >{\raggedright\arraybackslash}p{0.415\linewidth}
        >{\raggedright\arraybackslash}X@{}}
        \toprule[1.5pt]
        \textbf{Regime} & \textbf{Decision time and information} &
        \textbf{Effect scope and principal limitation} \\
        \midrule[1.2pt]
        \rowcolor{regimeLightBlue}
        \multicolumn{3}{c}{\textbf{\textit{Offline evolution}}} \\
        \addlinespace[1.5pt]
        Pre-deployment & Before deployment; training trajectories and
        outcomes when available ($\mathcal{I}^{\mathrm{off}}$) &
        Shared across target runs; cannot react to evidence revealed during a
        target run \\
        \addlinespace[3pt]
        \rowcolor{regimeLightBlue}
        \multicolumn{3}{c}{\textbf{\textit{Online evolution (deployment-time)}}} \\
        \addlinespace[1.5pt]
        Pre-run & At $t=0$; target instruction
        ($\mathcal{I}_{j,m}^{\mathrm{pre}}$) &
        Current run from $t=0$; cannot use evidence that emerges during
        execution \\
        \addlinespace[2pt]
        \rowcolor{regimeLightOrange}
        \textbf{Within-run} & At checkpoint $t_k$; instruction, partial
        trajectory, workspace, and prior edit status
        ($\mathcal{I}_{j,m}^{(k)}$), but not the eventual outcome &
        Remainder of the current run; lacks eventual outcome feedback and adds
        monitoring and intervention overhead \\
        \addlinespace[2pt]
        After-run\newline{\scriptsize same task} &
        After attempt $m$; completed trajectory and
        outcome when available ($\mathcal{I}_{j,m}^{\mathrm{post}}$) &
        Next attempt only; cannot rescue attempt $m$ and requires a restart \\
        \addlinespace[2pt]
        After-run\newline{\scriptsize task stream} &
        Before task $j$; earlier target trajectories
        and outcomes when available ($\mathcal{I}_{j}^{\mathrm{stream}}$) &
        Later tasks; cannot benefit the run that exposed the problem, and
        retained updates may overgeneralize or regress \\
        \addlinespace[2pt]
        \midrule[1.2pt]
        \rowcolor{regimeLightBrown}
        \textbf{\method (ours)} &
        \textcolor{regimeTeal}{\textbf{Offline:}} learn reusable
        $\mathcal{M}$ from $\mathcal{I}^{\mathrm{off}}$;
        \textcolor{regimeOrange}{\textbf{online, within-run:}} combine it with
        $\mathcal{I}_{j,m}^{(k)}$ without $y_{j,m}$ &
        \textcolor{regimeTeal}{Reusable $\mathcal{M}$ is carried across tasks;}
        \textcolor{regimeOrange}{each $\delta_{j,m}^{(k)}$ expires when its
        active run ends;} checkpoint supervision still adds online overhead \\
        \bottomrule[1.5pt]
    \end{tabularx}
    }
    \vspace{-3pt}
\end{table}
\section{Prompt and Output Templates}
\label{app:prompts}

This appendix presents the prompts for the online evolver and the offline \textsc{Reflect}, \textsc{Aggregate}, and \textsc{Update} stages.
Template variables and conditional blocks are populated at runtime.

\subsection{Prompt for Online Evolver}
\label{app:online-supervisor-prompt}
The following system prompt is used by the online evolver on Terminal-Bench 2.1.
The DeepSWE version uses the same checkpoint interface, intervention channels, and output format, but adds instructions for submitting committed Git changes and evaluating them in a separate container.
The Meta-Skill block is included only in configurations that use learned guidance.

\definecolor{promptblue}{HTML}{2F6690}
\definecolor{promptback}{HTML}{F1F7FC}

\newtcolorbox{promptcontentbox}[1]{
  enhanced,
  breakable,
  colback=promptback,
  colframe=promptblue,
  boxrule=0.8pt,
  arc=2pt,
  left=7pt,
  right=7pt,
  top=7pt,
  bottom=7pt,
  title={#1},
  fonttitle=\bfseries,
  coltitle=white,
  colbacktitle=promptblue,
  before skip=8pt,
  after skip=8pt
}

\begin{promptcontentbox}{Online Evolver System Prompt}
\begin{Verbatim}[fontsize=\small,breaklines=true,breakanywhere=true]
You are the supervisor of a long-horizon task.
Another agent — the *worker* — is doing the work: it operates in an isolated environment, decides its own actions, and holds its own context.
You do not act in that environment.
At each checkpoint you decide whether to change anything about the worker's situation, and you can only do so through the channels listed below.

The worker cannot see this conversation.
It perceives only what you send through a channel, and it treats what arrives as established.
Nothing you write as plain text reaches it.

## What arrives at each checkpoint

- The task statement, at the first checkpoint only.
- The outcome of your previous checkpoint: whether your edits loaded, or were rolled back, with the error.
- The worker's actions since the last checkpoint: its reasoning summaries, the commands it ran, and what the environment returned.
  Each block is truncated at 1,500 characters; the worker's full command outputs are on disk (see below).
- Two runner-computed sections at the end: **Repeated procedures** (commands the worker has re-run three or more times with small variations, with a fingerprint) and **Your channel budget** (notices used, persistent channels used).
- When the worker's last turn contained no tool call, a note saying it has stopped.
  Unless you send it back with `notice`, the state at that moment is final and is scored.
  The note says how many send-backs remain.

You are one continuous conversation for the whole task.
Your own history is the record of what you have sent and what came of it; `list_changes` lists the harness edits you have made.

## Read tools

`read_file`, `search_file_content`, `glob`, `list_directory`.
They reach the task directory (`/app`), the worker's harness (`/nexau-workspace`), the worker's saved command outputs (`/tmp/nexau_bash_tool_results/<id>/stdout.txt` and `stderr.txt`, the untruncated text of what the trace shows truncated), and the trial logs.
Other paths return an error.
They read; nothing you call executes in the worker's environment except a probe.

## Channels

Each channel's tool description carries its full contract: inputs, validation, limits, and what the worker will see.
What differs between channels is when the content reaches the worker, whether it persists, and what it guarantees.

| channel | reaches the worker | persists | guarantee |
|---|---|---|---|
| `notice` | at the start of its next turn, read once | no | advisory |
| `append_worker_prompt` / `rewrite_worker_prompt` | in its standing instructions, every turn from its next stretch of work | yes | advisory |
| `register_check` (Python, standard library, separate interpreter, reads files) | its output is appended to every command result | yes | reports |
| `register_probe` (bash, runs in the worker's environment, may execute programs) | its output is appended to every command result | yes | reports |
| `add_tool` | a new option in its action space | yes | optional |
| `add_skill` | a note it can load on demand | yes | optional |
| `add_middleware` | code around every model call or tool call | yes | enforced |
| `rollback` | restores the harness to its state at a checkpoint | — | — |

Limits enforced by the tools and the runner:

- One `notice` per checkpoint; a second call at the same checkpoint is refused.
  At most two notices per task while the worker is working; the tool refuses further ones.
  Sending a stopped worker back also uses `notice`; it is counted separately and its remaining number is shown in the stopped note.
- Harness edits (prompt, check, probe, tool, skill, middleware) take effect at the start of the worker's next stretch of work.
  If the rebuilt harness fails to load, the edit is rolled back and you are called again immediately with the error and one more attempt.
- After `add_tool`, `add_skill`, `add_middleware`, or a change to the standing instructions, the worker's next message states that the harness changed and what was added, including your description of it verbatim.
  Checks and probes are not announced; their output carries its own label on every command result.
- Aborting the worker is not available; the task runs to its own budget.
  A worker that is sent back keeps consuming that budget.

{% if knowledge_base_text %}
## Meta-Skill

Below is the whole Meta-Skill: entries learned offline from runs on *other* tasks of this benchmark by the same worker model.
It is the only precedent available to you, and it is reproduced here in full — there is no file to open and nothing further to read.
What it says about other tasks is precedent; anything you tell the worker about *this* task's files, outputs or environment must come from your own reads or from the trace.

Its entries for the opening moment are not reproduced here.
They are delivered once, in the first checkpoint's message, under the heading `# code_agent has just started`; that message stays in this conversation afterwards, so they remain available at every later checkpoint.

{{ knowledge_base_text }}
{% endif %}

## Output

Use the channel tools for anything you want to change.
End every checkpoint with exactly one `complete_task` call whose `result` is a JSON object:

```json
{"reasoning": "<for the log: what you saw, what you did or why nothing, and what backs each fact you sent>"}
```

`reasoning` never reaches the worker.
A notice written into `reasoning` is not delivered; only the `notice` tool delivers one.
\end{Verbatim}
\end{promptcontentbox}

\subsection{Shared System Prompt for Offline Stages}
\label{app:offline-system-prompt}

The \textsc{Reflect}, \textsc{Aggregate}, and \textsc{Update} stages use the same system prompt to define the task setting and the information available under conditions A, B, and C.

\begin{promptcontentbox}{Shared System Prompt}
\begin{Verbatim}[fontsize=\small,breaklines=true,breakanywhere=true]
<System>

A **worker** agent solves a software task in a git repository: it must change the repository so that a set of named tests newly passes without breaking the tests that already passed, and commit the change.
It never sees any of those tests.
Three settings are compared on the same tasks:

- **a**: the worker runs alone, unsegmented, with a simpler worker prompt (no probe/check middleware).

- **b**: a **supervisor** agent (the same model) watches the worker every few steps and may change the worker's harness for this task: send a one-off message (`notice`, at most two while the worker is working), add or rewrite standing instructions in the worker's prompt (`append_worker_prompt`, `rewrite_worker_prompt`), register a read-only probe or check whose output is appended to every worker command result, add a tool or skill, or roll back its own changes.
  When the worker stops, the supervisor accepts the result (the diff between the base commit and HEAD is submitted and graded; uncommitted work is discarded) or sends the worker back with a notice (at most twice).
  At the stop it also sees a runner-computed workspace state.
  It reads the **Meta-Skill** (entries with ids `E-NN`) and cites the ids it acts on.

- **c**: a supervisor that was given this task's acceptance conditions and verifier source at its first checkpoint, told to use them only to decide what is at risk and what to measure, and not to pass their values to the worker.
  It has no Meta-Skill.

Neither the worker nor the b supervisor ever sees the verifier.

</System>
\end{Verbatim}
\end{promptcontentbox}

\subsection{Prompt for \textsc{Reflect}}
\label{app:reflection-prompt}

The \textsc{Reflect} stage compares two groups of rollouts from the same training task and extracts candidate Meta-Skill entries.
The consolidated template below specifies the five contrast directions, entry fields, and output format.
Only the selected direction applies to each call, and privileged verifier information may explain training outcomes but cannot supply conditions or instructions unavailable to the online evolver.

\begin{promptcontentbox}{\textsc{Reflect} Prompt}
\begin{Verbatim}[fontsize=\small,breaklines=true,breakanywhere=true]
{system}

Your task is to compare two groups of runs of **one task** along the selected contrast direction and extract the experiences the Meta-Skill should carry from it.

<Selected Direction>
{direction_block}
The entry type for this call is `{type}`.
Apply only the corresponding direction below.
</Selected Direction>

<Contrast Directions>

1. informed — b runs that failed vs c runs that passed.
Type: act.

The left group is setting **b** runs that **failed**; the right group is setting **c** runs that **passed**.
The c supervisor knew the acceptance conditions.
Find what it did — which harness change, at which moment, on which evidence — that the b supervisor could also have done **without** the acceptance conditions.
The evidence must be something visible in the trace, the workspace, or the statement.
A lesson whose only basis is a value in the acceptance conditions is not allowed.
If the c supervisor's advantage was only that it knew a hidden value, the verdict is `no_supervisor_effect`.

2. effective — a runs that failed vs b runs that passed.
Type: act.

The left group is setting **a** runs that **failed**; the right group is setting **b** runs that **passed**.
Find a concrete intervention by the b supervisor that addresses a gap visible in the a runs: which harness change, at which moment, on which visible evidence.
The comparison provides directional evidence, not a checkpoint-level causal estimate.

**The two groups differ in more than the supervisor.**
Setting a uses a different worker directory: its system prompt says "your **only** tool", while setting b's says "your **main** tool ... other tools may be added during the task", and b's worker carries check/probe middleware.
Setting a runs **unsegmented**, while b runs in segments of a few turns; the effect of segmentation has not been measured on this benchmark.
**Do not credit the supervisor for differences that come from the worker scaffolding or from segmentation.**
Write an entry only when you can point at one concrete intervention in a b run and describe what the a runs did at the corresponding point without it.
If the b runs passed for reasons unrelated to the supervisor's actions, the verdict is `no_supervisor_effect`.

3. harmful — a runs that passed vs b runs that failed.
Type: avoid.

The left group is setting **a** runs that **passed**; the right group is setting **b** runs that **failed**.
Identify a concrete intervention associated with a subsequent failure or deviation, and compare it with the corresponding behavior in the passing a runs.
`avoid` concerns evidence of potential harm from a particular intervention, not merely an intervention with no observable effect.

**The a/b worker-scaffolding and segmentation differences described above apply here too.**
Do not blame the supervisor for a failure that the worker scaffolding or segmentation explains.
Write an entry only when you can identify one concrete intervention in a b run and show that the a runs, at the corresponding point, proceeded without the same problem.
Otherwise the verdict is `no_supervisor_effect`.

4. overreach — b runs that passed vs c runs that failed.
Type: avoid.

The left group is setting **b** runs that **passed**; the right group is setting **c** runs that **failed**.
The c supervisor knew more — the acceptance conditions — but its runs had worse outcomes.
Identify a concrete intervention associated with the failure: which channel, what content, at which moment.
Explain what visible evidence could have supported avoiding that intervention.
If the c failures are explained by the worker's own variation rather than by a supervisor action, the verdict is `no_supervisor_effect`.

5. unnecessary — a runs that passed vs b runs that passed with interventions.
Type: hold.

The left group is setting **a** runs that **passed** without any supervisor; the right group is setting **b** runs that **passed** and in which the supervisor **did intervene**.
Both groups succeeded, but this alone does not establish that an individual intervention was unnecessary.
The question is: **what observable situation appeared to require intervention, yet the worker could resolve it without supervision?**

You are also given two mechanical lists:
(1) In the a runs, every command that exited non-zero, with its failure category.
(2) In the b runs, the interventions whose content did not appear in the worker's later behavior.
**`when` should identify an observable feature of the kind in list (1)** — a failure class, a retry pattern, or a stage of the work.
`guidance` should say to hold and identify the evidence supporting that decision.
Do not write a `hold` entry for a situation where the a runs also show the worker getting stuck.

</Contrast Directions>

<Your Input>

- The task statement.
- The task's acceptance conditions and verifier source.
  The worker and the b supervisor never see these; use them to understand *why* a run passed or failed, never as content for an entry.
- Pass rates of the three settings on this task.
  These are independent samples, not matched pairs.
- A structured record of **every** run in the two groups: each worker turn's command, exit code, and the start of its output; each supervisor checkpoint at its true position, tagged `first` / `mid` / `final`, with the channel and text of what it did and the Meta-Skill ids it cited.
- The full text of the Meta-Skill entries cited in these runs.
  Only those entries are shown; the rest of the Meta-Skill is not provided.
{extra_input}
- Tools: `read_run(run, start, end)` returns turns T<start>..T<end> of one run with the supervisor checkpoints between them; `search_run(run, text)` returns matching lines.
  Use them when the record is too terse to judge a decision or when two runs appear to conflict.
  At most {tool_budget} calls.

</Your Input>

<Task Instructions>

1. Compare the two groups along the selected direction.
   **Judge each supervisor decision by what was visible to it at that moment, not by the terminal outcome.**
   Do not treat independent rollout comparisons as checkpoint-level counterfactuals.

2. Set the verdict to `entries` when the comparison supports a lesson for the b supervisor.
   Otherwise, use `no_supervisor_effect` and explain why: for example, the same failure appears on both sides, or the difference is explained by the worker's own variation.

3. If the verdict is `entries`, write one or more entries.
   **Type is fixed by the selected direction: `{type}`.
   Do not output it as an entry field.**
   Every entry carries a `time_point`.
   **One comparison may yield several entries at different time points.**
   An early constraint and a final verification requirement are different entries, even when they address the same failure.

4. For every cited Meta-Skill entry shown in the input, state whether it `helped`, `misled`, was `ignored_when_relevant`, or was `irrelevant` in these runs.

</Task Instructions>

<Time Point>

- `start` — the first checkpoint.
  The worker has taken one turn.
  Focus on before-acting decisions: preserving evidence that a routine first step would destroy, constraining how the worker probes, or excluding a route that available evidence shows cannot work.
  Do not give the worker the answer.

- `during` — the worker is working and there is an execution trace.
  Base decisions on observable evidence, such as a claim contradicted by command output, a recurring failure, or an unverified route exposed by the trace.

- `end` — the worker has stopped calling tools, the artifact exists, and workspace state has been collected.
  Assess whether the artifact and available validation support the worker's completion claim.

</Time Point>

<Fields of an Entry>

- `title`: 3 to 7 words naming the situation, not the action.

- `time_point`: start | during | end.

- `when`: one situation the b supervisor can recognize from what it actually holds at a checkpoint — the trace, a command's exit code, the workspace state, or the channel budget.
  Write it so that it can match a task from an unrelated domain.
  **Do not include a task name, domain noun (file format, algorithm, library, protocol, subject matter), path, command, version, or value.**
  Describe one situation, not a list of alternatives.
  Any wording specific to this task must be lifted one level up.

- `guidance`: specify both the harness action and, when applicable, what the message to the worker must carry.
  State which channel to use, at which moment, or what not to do.
  For `hold`, specify why the run should be left alone and what evidence supports that choice.
  When sending a message, distinguish between naming a concrete gap already exposed by the worker's trace and asking the worker to establish a property for itself.
  Explain which is appropriate.
  Do not reduce a specific, evidence-supported correction to a generic request for verification.

- `why`: describe what was observed in these runs, including the behavior following the intervention and the corresponding behavior without it.
  Do not claim causal certainty from independently sampled runs.

- `example`: one line starting with this task's name and a colon, describing concretely what happened.
  Do not include run labels or turn numbers.

- `runs`: the runs from which this entry was extracted.
  This field records provenance; no minimum is required.

- `derivable`: statement | worker_output | workspace_state | supervisor_read.
  State what evidence the b supervisor could use to make the decision.
  **A lesson that requires hidden acceptance conditions is not allowed.**

</Fields of an Entry>

<Output Format>

Output exactly one JSON object and nothing else.
Use English.

```json
{
  "analysis": "the comparison, with run and turn anchors",
  "verdict": "entries | no_supervisor_effect",
  "reason": "one or two sentences",
  "entries": [
    {
      "title": "",
      "time_point": "start | during | end",
      "when": "",
      "guidance": "",
      "why": "",
      "example": "<task>: ...",
      "runs": ["b2", "a4"],
      "derivable": "statement | worker_output | workspace_state | supervisor_read"
    }
  ],
  "entry_credit": [
    {
      "id": "E-03",
      "effect": "helped | misled | ignored_when_relevant | irrelevant",
      "runs": ["b1"],
      "note": "one sentence"
    }
  ],
  "recheck": "the raw-run reads you made and whether they changed anything; \"none\" if none"
}
```

Set `entries` to [] when the verdict is `no_supervisor_effect`.

</Output Format>
\end{Verbatim}
\end{promptcontentbox}

\subsection{Prompt for \textsc{Aggregate}}
\label{app:aggregation-prompt}

Each \textsc{Aggregate} call processes the candidates in one non-empty entry-type and time-point cell from the current training batch.
It uses the shared system prompt in Appendix~\ref{app:offline-system-prompt} and the instructions below.
Candidates are grouped by observable situation and intervention content, with each retained entry supported by at least two distinct training tasks.

\begin{promptcontentbox}{\textsc{Aggregate} Prompt}
\begin{Verbatim}[fontsize=\small,breaklines=true,breakanywhere=true]
You are assembling one cell of the Meta-Skill: **Type = `{type}`, Time point = `{time_point}`**.
Below are all candidate entries from this batch that fall in this cell.
Each candidate was written from **one** training task and is tagged with its source task and the comparison direction it came from.

<Time Point Meaning>

{time_point_def}

</Time Point Meaning>

<What to Produce>

Group the candidates into families and write one entry per family.

- **A family is defined by its `when`** — the situation in which the entry comes into play.
  Two candidates from different tasks belong to the same family when a supervisor would recognise the same situation, even when the tasks look nothing alike.

- **Hard constraint: every entry you output must draw on candidates from at least two different training tasks.**
  A candidate that cannot be joined with one from another task goes to `dropped` with a reason.
  Do not invent a family of one to hold it.

- Contradictory candidates belong in the same family, and the entry must say what separates them.
  "Intervene here" and "leave it alone here" describing the same situation is the most valuable thing in this material, not an error to resolve by dropping one side.

- **A family is one situation.**
  The merged `when` must still be one recognisable situation — at most two alternatives joined by "or".
  If candidates describe different situations (a false download, a missing companion tree and a partial aggregation are three situations, not one), they are different families — even if that leaves a family with one task, which is then dropped.
  Never fuse distinct situations into a list to satisfy the >=2-task constraint.

- There is no cap on the number of entries and no target; the >=2-task constraint and the one-situation rule are the only limits.

</What to Produce>

<Rules>

1. Nothing specific to a training task may appear in `title`, `when`, `guidance` or `why` — no task name, path, file name, field name, package, command, numeric value, and no domain noun that only fits one area.
   Those belong only in `examples`.

2. `why` must state the cross-task pattern: in what form it appeared on the different tasks.

3. `examples`: one line per source task, each starting with the task name and a colon (at most three).

4. `derivable`: the weakest basis among the merged candidates.

5. English.
   Each field one or two sentences.

6. Two candidates match when the situation *and* the kind of content the message carries are the same.
   "The supervisor named a gap the worker had already exposed" is one kind; "the supervisor asked the worker to establish a property" is another.
   Do not require the gap itself to be the same — that is domain-specific and belongs in `examples`.

</Rules>

<Output Format>

Output exactly one JSON object and nothing else.

```json
{
  "entries": [
    {
      "title": "",
      "when": "",
      "guidance": "",
      "why": "",
      "examples": ["<task>: ...", "<task>: ..."],
      "derivable": "statement | worker_output | workspace_state | supervisor_read",
      "from_candidates": ["C-03", "C-07"]
    }
  ],
  "dropped": [
    {
      "candidate": "C-05",
      "reason": "one sentence"
    }
  ]
}
```

`from_candidates` must list every candidate id in the family; the source tasks are computed from it.

</Output Format>

<Candidates>

{candidates}

</Candidates>
\end{Verbatim}
\end{promptcontentbox}

\subsection{Prompt for \textsc{Update}}
\label{app:update-prompt}

The \textsc{Update} stage is invoked once per training batch to incorporate aggregated entries into the current Meta-Skill.
It uses the shared system prompt in Appendix~\ref{app:offline-system-prompt} and the instructions below.
Updates consider both new entries and entry-level feedback, while preserving identifiers, source support, and examples.

\begin{promptcontentbox}{\textsc{Update} Prompt}
\begin{Verbatim}[fontsize=\small,breaklines=true,breakanywhere=true]
Your task is to decide how the Meta-Skill should change in response to the new entries from **one batch** of training tasks.
Choose one operation for each new entry, and add operations for the entry credit where the rules below require one.

<Your Input>

- The current Meta-Skill: every entry with its id and fields, plus its internal `support` (which training tasks it came from).
- The new entries from this batch (ids `C-NN`), each with `type`, `time_point`, fields and `support`.
- The entry credit collected in this batch: for cited entries, whether they helped, misled, were ignored when relevant, or were irrelevant, with runs.
- Tools `read_run(run, start, end)` and `search_run(run, text)` to confirm what a run shows before changing an existing entry.
  At most {tool_budget} calls.

</Your Input>

<Operations>

- **ADD**: the new entry describes a situation no existing entry covers.

- **MERGE**: the new entry and one existing entry (or two existing entries) have a `when` that lands on **the same observable**.
  The merged entry keeps the existing id; its `when` must still be **one recognisable situation** (at most two alternatives joined by "or") and its `guidance` **one action**; `support` becomes the union.
  Give the id, optionally `absorb` (other existing ids folded in), and only the fields that change.

- **REFINE**: an existing entry covers the **same situation**; improve it without widening it — sharpen or qualify the guidance, correct the why, change the type if the evidence shows it, or append an example.
  **Never add a triggering condition.**
  Give the id and only the fields that change.

- **DELETE**: see rule 4.

- **NONE**: already covered, or adds nothing.

</Operations>

<Rules>

1. Before calling a new entry "different from every existing one" (ADD), check whether it lands on the same observable as an existing entry — if it does, that is a MERGE, not an ADD.

2. REFINE never widens `when`.
   To widen, use MERGE, and only if the result is still one recognisable situation.
   If the merged `when` would become a list of unrelated observables, or the merged `guidance` would need "depending on the case" to branch, **it is not a merge: keep two entries.**

3. An entry judged `ignored_when_relevant` gets at most an appended example, not a new `when`.

4. **DELETE** only when an entry was judged `misled` in at least two runs of this batch **and** no qualifier can be written that narrows it to a situation where it is still useful.
   Otherwise add the qualifier (REFINE).

5. Examples are only appended (at most three per entry); existing examples are never rewritten.

6. Cross-task wording in title, when and guidance: no task name, domain noun, path, command, version or value.

7. Call the one-off message channel **notice**.

8. Every operation other than NONE cites at least two runs as `<task>:<run>` (for example `dna-insert:b2`).

9. Length targets: title 60 characters, when 240, guidance 360, why 300, example 260.

</Rules>

<Output Format>

Output exactly one JSON object and nothing else.
Use English.

```json
{
  "operations": [
    {
      "op": "ADD | MERGE | REFINE | DELETE | NONE",
      "candidate": "C-NN or null",
      "id": "existing id for MERGE / REFINE / DELETE, else null",
      "absorb": ["other existing ids, MERGE only"],
      "entry": {
        "title": "",
        "when": "",
        "guidance": "",
        "why": "",
        "type": ""
      },
      "add_example": "for REFINE only, or null",
      "runs": ["<task>:b1", "<task>:a3"],
      "reason": "one sentence"
    }
  ],
  "recheck": "the raw-run reads you made and whether they changed anything; \"none\" if none"
}
```

For MERGE and REFINE, `entry` holds only the fields that change.
For DELETE and NONE, `entry` may be empty.

</Output Format>

<Current Meta-Skill>

{kb}

</Current Meta-Skill>

<New Entries from This Batch>

{candidates}

</New Entries from This Batch>

<Entry Credit from This Batch>

{credit}

</Entry Credit from This Batch>
\end{Verbatim}
\end{promptcontentbox}
\section{Meta-Skill Structure and Content}
\label{app:meta-skill-entries}

\subsection{Entry Structure}
\label{app:meta-skill-entry}

A Meta-Skill is a versioned textual library that guides harness evolution at checkpoints.
Each entry associates an observable situation with guidance on whether and how to intervene.
The evolver interprets these entries using evidence from the active run to select a task-local harness edit or leave the harness unchanged.
We denote the frozen library by $\mathcal{M}=\{e_i\}_{i=1}^{N}$, with the entry fields summarized in Table~\ref{tab:meta-skill-entry-content}.

\begin{table}[H]
  \centering
  \caption{\textbf{Fields of a Meta-Skill entry.}
  Each entry specifies an observable situation, evolution guidance, and supporting training examples.}
  \label{tab:meta-skill-entry-content}
  \small
  \renewcommand{\arraystretch}{1.15}
  \setlength{\tabcolsep}{6pt}
  \begin{tabularx}{0.96\linewidth}{
    @{\hspace{5pt}}
    >{\raggedright\arraybackslash}p{0.23\linewidth}
    >{\raggedright\arraybackslash}X
    @{\hspace{5pt}}
  }
    \toprule
    \textbf{Field} & \textbf{Content} \\
    \midrule
    \texttt{ID / title}
    & A stable identifier for citation and a short title describing the situation. \\
    \texttt{Type}
    & \texttt{act} supports intervention; \texttt{hold} supports leaving the run unchanged; \texttt{avoid} discourages a particular intervention. \\
    \texttt{Time point}
    & \texttt{start}, \texttt{during}, or \texttt{end}, indicating the execution stage to which the entry applies. \\
    \texttt{When to refer}
    & An observable condition that the evolver can assess from the task instruction, trajectory, permitted workspace observations, or channel budget. \\
    \texttt{Guidance}
    & The harness action to take or avoid, including the channel and message content when applicable. \\
    \texttt{Why}
    & The pattern observed across training tasks that supports the guidance, including relevant qualifications. \\
    \texttt{Reference examples}
    & Up to three examples from distinct training tasks illustrating the situation and observed behavior. \\
    \bottomrule
  \end{tabularx}
\end{table}

The \texttt{When to refer} field describes an observable condition at a level that permits reuse across tasks.
Vague judgments such as ``the approach looks wrong'' provide little basis for matching an entry to checkpoint evidence.
Conversely, exact file names, commands, packages, or expected answers tie a trigger to a particular instance.
We therefore express triggers through observable relations, such as a claim contradicted by command output, repeated attempts that produce no new evidence, or an artifact inconsistent with the task agent's completion claim.
Task-specific details are confined to reference examples and internal provenance records.
Construction metadata, including supporting task identifiers and candidate provenance, is retained separately to support aggregation and subsequent updates.

\FloatBarrier
\subsection{Content of Meta-Skills}
\label{app:meta-skill-content}

Entries are organized by type and execution stage.
The \texttt{act} entries describe situations in which training contrasts support an intervention, such as an unverified execution route, an unresolved contradiction, or an unsupported completion claim.
The \texttt{hold} entries identify situations where the task agent is making progress or recovering without assistance.
The \texttt{avoid} entries describe interventions associated with adverse outcomes, such as prematurely selecting an interpretation or redirecting a search that is still advancing.
These types guide different aspects of a checkpoint decision: \texttt{act} supports a change, \texttt{hold} supports continuation, and \texttt{avoid} excludes a particular change while leaving other interventions available.

The execution-stage labels further specify when the guidance applies.
Start entries concern early constraints, evidence preservation, and route selection.
During entries concern evidence revealed as work proceeds, including progress, contradictions, and unverified behavior.
End entries concern whether the delivered artifact and available validation support acceptance.

Cross-task aggregation retains reusable guidance supported by at least two distinct training tasks.
Raw verifier text, hidden expected outputs, task solutions, and held-out information are excluded from the deployed guidance.
At deployment, the frozen Meta-Skill is supplied directly in the evolver's conversation.
Entries labeled \texttt{start} are delivered in the first checkpoint message, while the remaining entries are included in the system prompt.
All entries remain available in the evolver's conversation at subsequent checkpoints.
The evolver records the entry IDs it uses, or indicates that no entry applies, and grounds each decision in evidence from the active run.
Checkpoint decisions, task-local edits, and test-time outcomes do not update the frozen library.

\definecolor{metaskillblue}{HTML}{2F6690}
\definecolor{metaskillback}{HTML}{F1F7FC}
\newtcolorbox{metaskillcontentbox1}{
  enhanced,
  breakable,
  colback=metaskillback,
  colframe=metaskillblue,
  boxrule=0.8pt,
  arc=2pt,
  left=7pt,
  right=7pt,
  top=7pt,
  bottom=7pt,
  title={Meta-Skills for Terminal-Bench 2.1},
  fonttitle=\bfseries,
  coltitle=white,
  colbacktitle=metaskillblue,
  before skip=8pt,
  after skip=8pt
}

\newtcolorbox{metaskillcontentbox2}{
  enhanced,
  breakable,
  colback=metaskillback,
  colframe=metaskillblue,
  boxrule=0.8pt,
  arc=2pt,
  left=7pt,
  right=7pt,
  top=7pt,
  bottom=7pt,
  title={Meta-Skills for DeepSWE 1.1},
  fonttitle=\bfseries,
  coltitle=white,
  colbacktitle=metaskillblue,
  before skip=8pt,
  after skip=8pt
}

\begin{metaskillcontentbox1}
  Entries learned offline from supervised and unsupervised runs on \textbf{other} tasks of this benchmark by the same worker model.
What they say about those tasks is precedent, not a fact about the task in front of you: anything you tell the worker about \emph{this} task's files, outputs, or environment must come from your own reads or from the trace.

Every action available to you changes this run's harness; the actions differ in how long the change lasts.
A \texttt{notice} reaches one worker turn.
A standing instruction (\texttt{append\_worker\_prompt} or \texttt{rewrite\_worker\_prompt}), a check, and a probe reach every turn from the worker's next stretch of work onward.
Tools, skills, and middleware change the worker's structure.

\paragraph{How to read an entry.}
\begin{itemize}[leftmargin=*,nosep]
  \item \textbf{Type}---\texttt{act}: a situation where a harness change helped; \texttt{hold}: a situation that looks like it needs intervention and does not, so leave the run alone; \texttt{avoid}: an intervention that did harm.
  \item \textbf{Time point}---\texttt{start} (first checkpoint), \texttt{during} (worker working), or \texttt{end} (worker stopped).
  \item \textbf{When to refer}---the observable situation that brings the entry into play.
  Match on this.
  \item \textbf{Guidance}---what to do, or what not to do.
  \item \textbf{Why}---what was observed in the training runs.
  \item \textbf{Reference examples}---how it looked on training tasks; illustration only.
\end{itemize}

An entry whose situation you do not see does not apply.
Nothing here ranks the entries or states how often each was drawn on.
Write in \texttt{reasoning} the IDs of the entries you act on and the IDs of entries whose situation you saw but decided against.
If none applies, say so.

\subsubsection*{At the first checkpoint}

The worker has taken one turn.
These entries are for this moment only: a standing instruction installed now reaches every one of its remaining turns, and the route it will follow is not yet fixed.
Match on the situation as with any other entry.

\paragraph{E-02---Externally observable end state named.}
\textbf{Type.} \texttt{act}

\textbf{Time point.} \texttt{start}

\textbf{When to refer.} The statement requires an end state that can only be confirmed by observing the system from outside it, rather than by a file or a build result.

\textbf{Guidance.} Install a standing instruction that completion requires a fresh observation of that external view taken before and after the action, under the same conditions.
Do not let a reachable channel, a running process or a returned status stand in for it.

\textbf{Why.} Supervisors that established this expectation at the first checkpoint obtained a visible observed change and passed, while those that concentrated on channel availability accepted a reported status without a comparable observation.

\textbf{Reference examples.} (1) \texttt{install-windows-3.11}: passing runs compared captures of the external view around injected input, while failed runs confirmed only that the channel accepted commands.

\paragraph{E-03---Obvious first move destroys or under-covers.}
\textbf{Type.} \texttt{act}

\textbf{Time point.} \texttt{start}

\textbf{When to refer.} The most direct way into the task would either alter something that cannot be recovered once altered, or would cover less than the requirement actually asks for---and the worker has not yet committed to a route.

\textbf{Guidance.} Install a standing instruction that constrains the method before it acts: what must be preserved or worked on only in copy, how wide the search or enumeration has to be, and which behaviour the requirement implies that its wording does not spell out.
Say nothing about what the answer is.

\textbf{Why.} On these tasks the deciding step was the first one.
Where the constraint was installed the work survived; where the first ordinary action went ahead unconstrained, the evidence or the coverage was already gone by the time anything else could be said.

\textbf{Reference examples.} (1) \texttt{sqlite-db-truncate}: the artifact that had to be recovered was also the thing an ordinary read would rewrite.
(2) \texttt{dna-assembly}: the established ordering and boundaries had to be preserved through an encoding step that silently reorders them.

\paragraph{E-01---Interpretation asserted before inspection.}
\textbf{Type.} \texttt{avoid}

\textbf{Time point.} \texttt{start}

\textbf{When to refer.} At the first checkpoint the worker has only listed the inputs it was given and has not yet displayed or compared their contents.

\textbf{Guidance.} Do not use a notice or a standing instruction to fix an interpretation, a boundary or a transformation of material the worker has not yet inspected.
A standing instruction may constrain how it examines the material and what it must preserve; leave the reading of the material to it.

\textbf{Why.} A first-checkpoint message that supplied a specific interpretation was adopted verbatim and produced an internally consistent but wrong result, while the unsupervised run made its own comparison and read the material differently.

\textbf{Reference examples.} (1) \texttt{dna-insert}: a first-checkpoint notice stated a particular insertion boundary before inspection; the worker adopted it, while the passing unsupervised run found a different alignment itself.

\subsubsection*{While the worker is working}

\paragraph{E-07---Reported success with unchanged observation.}
\textbf{Type.} \texttt{act}

\textbf{Time point.} \texttt{during}

\textbf{When to refer.} An action reports success, and a later direct observation of the very thing it was meant to change shows the same state as before.

\textbf{Guidance.} Send one notice stating that the reported success is contradicted by the observation, and require a fresh comparison under stable conditions before the worker builds anything further on the claim.

\textbf{Why.} Runs that treated an accepted command as proof of effect accepted a state that had not changed; runs that treated the unchanged observation as unresolved kept working until the action visibly took effect.

\textbf{Reference examples.} (1) \texttt{install-windows-3.11}: repeated accepted inputs left the external view unchanged in failed runs, while passing runs continued until an input visibly changed it.

\paragraph{E-08---Restriction the evidence has not established.}
\textbf{Type.} \texttt{act}

\textbf{Time point.} \texttt{during}

\textbf{When to refer.} The worker narrows the problem by a constraint that neither the requirements nor any observed failure supports---coupling things that vary independently, or rejecting inputs that were accepted before---and that narrowing is what makes its search come up empty or its behaviour change.

\textbf{Guidance.} Send one notice to drop the unsupported restriction, keep only what the requirements and the observed failures establish, and retry.
Do not accept an empty result under an unproven restriction as evidence that none exists.

\textbf{Why.} An assumed coupling eliminated the viable candidates entirely; an inferred rejection rule changed behaviour outside the failure that was actually demonstrated.
Both turned an unproven limitation into a requirement.

\textbf{Reference examples.} (1) \texttt{dna-insert}: forcing two independently variable components to the same setting returned nothing, and enumerating them separately produced a working combination at once.
(2) \texttt{fix-code-vulnerability}: a message adding rejection of a further boundary input produced a failing artifact, while passing repairs restricted themselves to the demonstrated unsafe class.

\paragraph{E-09---Comparison the worker ran reports a mismatch.}
\textbf{Type.} \texttt{act}

\textbf{Time point.} \texttt{during}

\textbf{When to refer.} A comparison or check the worker itself ran reports that its result differs from a reference in membership, count, or value, and later turns proceed without that difference having been resolved.

\textbf{Guidance.} Send one notice requiring the worker to isolate the first difference, decide whether any permitted normalisation explains it, repair otherwise, and rerun that same comparison.
Do not let a narrower check that passes stand in for the one that failed.

\textbf{Why.} Supervisors that treated a failing self-run comparison as blocking obtained corrected and revalidated artifacts; runs that let it pass continued on an artifact whose own evidence contradicted it.

\textbf{Reference examples.} (1) \texttt{regex-chess}: a randomised comparison exposed a discrepancy after a narrower check had passed, and passing runs required it reconciled and repeated.
(2) \texttt{dna-assembly}: a reported failing round-trip check was replaced by another hand-built calculation instead of being resolved.

\paragraph{E-11---Located material bypassed for a surrogate.}
\textbf{Type.} \texttt{act}

\textbf{Time point.} \texttt{during}

\textbf{When to refer.} The trace shows the worker has already located the thing the requirement names---the material itself, or the authority it must come from---and it is nonetheless working from something it found separately.

\textbf{Guidance.} Send one notice to open or cross-check against what it already located before going further, and to state which of the two its result rests on.
Do not let a calculation from the surrogate become the final evidence.

\textbf{Why.} Workers that kept working from a surrogate after having exposed the named source delivered results the named source did not support; redirection to what they had already found produced passing results.

\textbf{Reference examples.} (1) \texttt{mteb-leaderboard}: the official source had been exposed but the answer was computed from a separately obtained copy, and passing runs followed the named source instead.
(2) \texttt{build-pov-ray}: the listed material containing what was needed was passed over for continued speculative search until a message pointed back at it.

\paragraph{E-04---Isolated environment failure with visible reformulation.}
\textbf{Type.} \texttt{hold}

\textbf{Time point.} \texttt{during}

\textbf{When to refer.} The worker has already begun substantive work on the deliverable, and a single command then exits non-zero because the execution environment rejects an invocation feature or an auxiliary capability is unavailable, with the next visible action reformulating the same work.

\textbf{Guidance.} Hold: do not supply a replacement or spend a notice on that failure.
This concerns reacting to the failure itself---it does not bar a method, scope, or acceptance constraint you would have installed anyway.
Intervene once the same class of failure recurs, or once a substantial share of the time budget is gone with no progress past it.

\textbf{Why.} Across many training tasks this signature appeared in runs that passed with no supervisor at all; the worker removed the unsupported feature, installed the capability, or switched route on its own turn.
Messages spent here changed nothing the worker was not already doing.

\textbf{Reference examples.} (1) \texttt{sqlite-with-gcov}: an early command was rejected over an unsupported control option, then retried in a compatible form and the build completed.
(2) \texttt{financial-document-processor}: an unavailable inspection utility was followed by capability discovery and a successful alternative.
(3) \texttt{reshard-c4-data}: one unsupported interpreter option failed, then the same test ran through a compatible interpreter.

\paragraph{E-05---Failure output carries its own correction.}
\textbf{Type.} \texttt{hold}

\textbf{Time point.} \texttt{during}

\textbf{When to refer.} The worker has already begun substantive work, a command fails, and its own output already names what to fix---the location it actually found, the accepted alternative, or the required form---and the next visible action reads that output rather than repeating the command unchanged.

\textbf{Guidance.} Hold: do not name the fix and do not register a check for it.
The output has bounded the next experiment and the worker is already reading it.
This concerns reacting to that failure---it does not bar a method or acceptance constraint you would have installed anyway.
Intervene if the corrected attempt fails the same way again.

\textbf{Why.} On separate tasks the worker recovered from this signature within one or two turns without help, and messages that named the fix arrived after the worker had already applied it.

\textbf{Reference examples.} (1) \texttt{crack-7z-hash}: the command reported an item missing after its own output had listed that item at a deeper location, and the worker used the listed location next.
(2) \texttt{mteb-retrieve}: a rejected identifier came with a compatible alternative in the error, which the worker applied before resolving the next interface error.
(3) \texttt{regex-log}: a self-written check failed on an assertion and the worker printed the actual result instead of rerunning the check blindly.

\paragraph{E-06---Route replaced while exploration still advancing.}
\textbf{Type.} \texttt{avoid}

\textbf{Time point.} \texttt{during}

\textbf{When to refer.} One attempt along the worker's current route fails or returns nothing usable, while the surrounding turns show it varying its approach, following new leads, or not yet having tried the ordinary way of driving the thing it just built.

\textbf{Guidance.} Do not redirect the route, replace the approach, or rewrite standing instructions on that evidence.
Let the ordinary next interaction happen or let the varied exploration continue.
Act once repeated attempts add nothing, or once a substantial share of the task's time budget is gone and the exploration has not narrowed.

\textbf{Why.} Interventions on this evidence moved workers off routes that were about to succeed and onto more fragile ones that then failed, while unsupervised runs meeting the same signal proceeded and passed.

\textbf{Reference examples.} (1) \texttt{qemu-startup}: a passive connection closed with no output, and a redesign message led to a fragile replacement while another run simply sent the ordinary initiating input and kept the simple service.
(2) \texttt{mteb-leaderboard}: a redirection away from a fallback that was producing leads ended in the wrong answer, while runs that continued through the same failure reached the right one.
(3) \texttt{build-pov-ray}: a suggested alternate source also failed, and continued independent searching found the working route.

\paragraph{E-10---Ambiguity narrowed before the evidence resolves it.}
\textbf{Type.} \texttt{avoid}

\textbf{Time point.} \texttt{during}

\textbf{When to refer.} The worker's own output shows several candidate results produced by different readings of the requirement, and nothing it has gathered excludes any of them.

\textbf{Guidance.} Do not send a notice that favours one candidate.
If you act at all, use a notice that requires the worker to settle which reading the requirement calls for, from the material it can read, before it picks one.

\textbf{Why.} A message emphasising one candidate was followed by delivery of that candidate alone, while unsupervised runs that kept the ambiguity open investigated further and delivered the complete result.

\textbf{Reference examples.} (1) \texttt{count-dataset-tokens}: a message emphasising one component was followed by an answer containing only that component's subtotal, while passing unsupervised runs combined both.

\subsubsection*{When the worker has stopped}

\paragraph{E-12---Completion shown only indirectly.}
\textbf{Type.} \texttt{act}

\textbf{Time point.} \texttt{end}

\textbf{When to refer.} The worker has stopped, and the only validation in the trace either discards the output, reuses values it asserted by hand, or comes from something other than the artifact it delivered---the delivered thing has not itself been run and inspected.

\textbf{Guidance.} Send it back to run the delivered artifact end to end under ordinary conditions and inspect what comes out.
Accept only after the claimed result has been derived from the delivered thing itself.

\textbf{Why.} Across several tasks a stopped worker presented a check that succeeded while never exercising what it delivered; acceptance failed until the artifact itself was run and its output read.

\textbf{Reference examples.} (1) \texttt{build-pov-ray}: a check that discarded output did not show that the ordinary output path produced a readable result.
(2) \texttt{dna-insert}: a reported match was computed from assumed intermediates rather than by recomposing the delivered components.
(3) \texttt{mteb-leaderboard}: a correctly formed answer file rested only on a substitute source.

\paragraph{E-13---Delivered artifact contradicts its own trace.}
\textbf{Type.} \texttt{act}

\textbf{Time point.} \texttt{end}

\textbf{When to refer.} The delivered artifact carries a value, a direction, or an implementation choice that differs from something the worker's own earlier output showed, or from an explicit rule in the requirements.

\textbf{Guidance.} Do not accept on surface form alone.
Send it back to reconcile the delivered choice against that earlier evidence or stated rule, regenerate whatever depends on it, and inspect the revised artifact.

\textbf{Why.} Superficially complete artifacts kept a choice their own trace or the stated rule contradicted, and the supervisors accepted them; reconciling the visible discrepancy before acceptance corrected the result.

\textbf{Reference examples.} (1) \texttt{count-dataset-tokens}: a plain result file held one measured component although the trace had measured its companion too.
(2) \texttt{bn-fit-modify}: the delivered relations encoded the direction opposite to the stated rule, and the worker's own explanation said so.
(3) \texttt{fix-code-vulnerability}: the final source added an error for an input the original had returned unchanged.

\paragraph{E-14---Stated exceptional behaviour never checked.}
\textbf{Type.} \texttt{act}

\textbf{Time point.} \texttt{end}

\textbf{When to refer.} The requirements call out exceptional, boundary, or previously accepted behaviour, and the trace shows validation only along the routine path.

\textbf{Guidance.} Send it back to exercise those stated cases directly, repair any difference, then rerun the routine validation before accepting.

\textbf{Why.} Routine success left material exceptional behaviour unproved, and in one case a change silently narrowed behaviour that had been accepted before; focused checks either found the defect or established that it was absent.

\textbf{Reference examples.} (1) \texttt{regex-chess}: focused comparisons of the stated exceptional cases found defects the routine validation had passed over.
(2) \texttt{fix-code-vulnerability}: the delivered change rejected an input the original accepted, which the main suite did not catch.
\end{metaskillcontentbox1}

\begin{metaskillcontentbox2}
  Entries learned offline from supervised and unsupervised runs on \textbf{other} tasks of this benchmark by the same worker model.
What they say about those tasks is precedent, not a fact about the task in front of you: anything you tell the worker about \emph{this} task's files, outputs, or environment must come from your own reads or from the trace.

Every action available to you changes this run's harness; the actions differ in how long the change lasts.
A \texttt{notice} reaches one worker turn.
A standing instruction (\texttt{append\_worker\_prompt} or \texttt{rewrite\_worker\_prompt}), a check, and a probe reach every turn from the worker's next stretch of work onward.
Tools, skills, and middleware change the worker's structure.

\paragraph{How to read an entry.}
\begin{itemize}[leftmargin=*,nosep]
  \item \textbf{Type}---\texttt{act}: a situation where a harness change helped; \texttt{hold}: a situation that looks like it needs intervention and does not, so leave the run alone; \texttt{avoid}: an intervention that did harm.
  \item \textbf{Time point}---\texttt{start} (first checkpoint), \texttt{during} (worker working), or \texttt{end} (worker stopped).
  \item \textbf{When to refer}---the observable situation that brings the entry into play.
  Match on this.
  \item \textbf{Guidance}---what to do, or what not to do.
  \item \textbf{Why}---what was observed in the training runs.
  \item \textbf{Reference examples}---how it looked on training tasks; illustration only.
\end{itemize}

An entry whose situation you do not see does not apply.
Nothing here ranks the entries or states how often each was drawn on.
Write in \texttt{reasoning} the IDs of the entries you act on and the IDs of entries whose situation you saw but decided against.
If none applies, say so.

\subsubsection*{At the first checkpoint}

The worker has taken one turn.
These entries are for this moment only: a standing instruction installed now reaches every one of its remaining turns, and the route it will follow is not yet fixed.
Match on the situation as with any other entry.

\paragraph{E-07---Do not repeat an explicit setup prerequisite during initial reconnaissance.}
\textbf{Type.} \texttt{hold}

\textbf{Time point.} \texttt{start}

\textbf{When to refer.} At the first checkpoint, the worker has only performed normal repository reconnaissance and an explicit routine setup prerequisite remains incomplete.

\textbf{Guidance.} Hold: do not spend a notice restating the prerequisite before the worker has begun implementation.
Initial reconnaissance is compatible with completing an explicit setup step next, so preserve the notice unless progress later contradicts that expectation.

\textbf{Why.} Across tasks, workers remained in the initial repository state while inspecting it, then performed the stated setup step as their next substantive action; a supervisory reminder did not produce a distinct recovery or outcome.

\textbf{Reference examples.} (1) \texttt{geo-shapeindex-serialization}: workers inspected the repository and then created the required dedicated branch without needing a reminder.
(2) \texttt{ofetch-per-origin-circuit-breaker}: the worker reported the default branch during reconnaissance, received a reminder, and then made the same routine branch transition seen in unassisted runs.

\subsubsection*{While the worker is working}

\paragraph{E-01---Audit scoped state across execution and recovery branches.}
\textbf{Type.} \texttt{act}

\textbf{Time point.} \texttt{during}

\textbf{When to refer.} A change introduces or routes through scoped execution state, and a child or caller branch or alternate completion path either fails to receive required state or receives state outside its permitted scope.

\textbf{Guidance.} Send a targeted notice to trace state ownership, lifetime, and every branch or boundary that creates, inherits, establishes, or recovers the state.
Require focused checks for intended and excluded paths before finalization.

\textbf{Why.} State placed only for the ordinary caller or normal return can be unavailable to an allowed alternate path, while incorrect propagation can also leak it into an excluded path.

\textbf{Reference examples.} (1) \texttt{abs-module-cache-flags}: a nested module environment did not retain caller-held loader settings.
(2) \texttt{mobly-grouped-test-barriers}: shared lifecycle handling established protected context for excluded paths, while another ordinary branch lacked required phase state.
(3) \texttt{awilix-async-container-initialization}: an inherited definition was owned by an ancestor while lifecycle status was consulted locally.

\paragraph{E-02---Directly exercise an unproven changed behavior route.}
\textbf{Type.} \texttt{act}

\textbf{Time point.} \texttt{during}

\textbf{When to refer.} The worker's own trace has surfaced a second route for the behaviour it just changed---it named, opened, or ran something on that route---and it has moved on to other work without running that route's behaviour even once.

\textbf{Guidance.} Send one targeted notice about that route.
When the worker's own trace already shows what is missing on it---a call that does not forward a setting, a branch that was never wired, or a sibling left at the old behaviour---name that gap and let the worker repair it; asking only for a check leaves the worker to rediscover what the trace already exposed.
When the trace shows no such gap, ask instead for a focused behavioural check of that route covering representative success and failure, before relying on broad regression results.

\textbf{Why.} Generic suites, adjacent-route checks, or stale focused tests can appear reassuring while leaving changed behavior unexercised; distinct sibling handling can then retain a defect that only direct behavioral checks expose.

\textbf{Reference examples.} (1) \texttt{numba-stencil-boundary-modes}: ordinary behavior was updated while the parallel transformation needed corresponding propagation.
(2) \texttt{valibot-recursive-schema-composition}: a synchronous repair left an asynchronous sibling closure-bound, and an async wrapper delegated to synchronous replacement behavior.
(3) \texttt{awilix-async-container-initialization}: coverage exercised one injection mode while another mode inferred dependencies differently.

\paragraph{E-03---Escalate incomplete compatibility parity evidence.}
\textbf{Type.} \texttt{act}

\textbf{Time point.} \texttt{during}

\textbf{When to refer.} A compatibility-sensitive change either differs from established behavior in available evidence or has been checked against only a subset of the required behavior classes.

\textbf{Guidance.} Send a notice to create focused differential checks against the established behavior across the unexercised cases, repair every observed difference, and rerun the comparisons before delivery.

\textbf{Why.} One task exposed a regression after a broad rewrite, while another showed replay-order differences and initially covered only a narrow portion of its compatibility surface.

\textbf{Reference examples.} (1) \texttt{abs-module-cache-flags}: an existing import behavior failed after a loading rewrite and needed restoration.
(2) \texttt{scc-bounded-memory-spilling}: bounded replay differed from established output ordering, and broader comparisons later exposed additional output differences.

\paragraph{E-04---Audit new capability wiring to its public route.}
\textbf{Type.} \texttt{act}

\textbf{Time point.} \texttt{during}

\textbf{When to refer.} A newly introduced control or capability is present in implementation but cannot traverse its owning registration or intended public route.

\textbf{Guidance.} Send a targeted notice to trace the capability from its owning registration through every public entry and receiving handler.
Require missing defaults, admission, bindings, or routes to be repaired, then directly exercise each affected public route.

\textbf{Why.} Implementation can be locally complete while omitted registration, admission, handler wiring, or public exposure prevents the capability from being usable end to end.

\textbf{Reference examples.} (1) \texttt{onedump-dump-encryption-pipeline}: direct source loading returned no result when an unrelated enabled-state check was false.
(2) \texttt{valibot-recursive-schema-composition}: the implementation was available through one aggregate but absent from the required public-facing aggregate.
(3) \texttt{arcane-drift-detection-baselines}: newly used controls were absent from the settings model and default configuration.

\paragraph{E-08---Preserve established write boundaries.}
\textbf{Type.} \texttt{act}

\textbf{Time point.} \texttt{during}

\textbf{When to refer.} A changed writer performs some externally visible writes outside an established scoped or atomic boundary.

\textbf{Guidance.} Send a targeted notice to inspect the complete write and publication sequence, restore the established boundary for ordinary behavior, and keep new behavior appropriately scoped.
Require a focused compatibility or concurrent-observation check before delivery.

\textbf{Why.} Writes outside the established boundary can change existing visibility guarantees or expose observers to partial or mixed results.

\textbf{Reference examples.} (1) \texttt{sqlite-utils-safe-import-checkpoints}: direct writes bypassed the prior connection-context behavior and changed caller-visible persistence.
(2) \texttt{vulture-persistent-analysis-cache}: only the primary artifact was safely replaced while companion artifacts were written directly.

\paragraph{E-09---Do not redirect solely for unavailable verification prerequisites.}
\textbf{Type.} \texttt{hold}

\textbf{Time point.} \texttt{during}

\textbf{When to refer.} A verification attempt explicitly reports that an external dependency or executable is unavailable and therefore cannot assess project behavior.

\textbf{Guidance.} Hold: do not spend a notice on setup, replacement tooling, or implementation redirection solely for this failure.
Leave the worker to use available focused checks, source review, or other substantive validation unless later evidence indicates a behavioral defect.

\textbf{Why.} Unavailable verification prerequisites can block broad or ancillary checks without indicating an implementation problem, while available validation may still support progress.

\textbf{Reference examples.} (1) \texttt{cattrs-partial-structuring-recovery}: broad verification stopped at unavailable imports, and quality-tool attempts reported missing executables or failed downloads; focused checks and review still completed the change.
(2) \texttt{vulture-persistent-analysis-cache}: an unavailable ancillary quality command was followed by available repository checks and a successful committed implementation.

\subsubsection*{When the worker has stopped}

\paragraph{E-05---Reopen delivery when concrete evidence refutes completion.}
\textbf{Type.} \texttt{act}

\textbf{Time point.} \texttt{end}

\textbf{When to refer.} The stopped artifact or its trace contains concrete evidence that a claimed behavior remains wrong, or a corrective edit for it is still outside the commit.

\textbf{Guidance.} Do not accept.
Name the specific contradiction the evidence shows---which claimed behaviour, and which observation refutes it---and send the worker back to repair exactly that and commit or amend it.
Keep the repair scoped to the contradiction: at this point the worker has little budget left to establish that a wider rewrite broke nothing, and a change that reaches beyond the named gap can fail behaviour that already worked.

\textbf{Why.} Across tasks, final review exposed either visibly incorrect behavior, an implementation form that defeated the claimed guarantee, or a repair stranded after interrupted validation; each meant the delivered commit did not support the completion claim.

\textbf{Reference examples.} (1) \texttt{abs-module-cache-flags}: a resolver repair remained modified but uncommitted after a failing validation-and-amend chain stopped.
(2) \texttt{kysely-window-grouping-helpers}: final manual output still showed invalid bound ordering despite the committed completion claim.
(3) \texttt{valibot-recursive-schema-composition}: the committed type-level implementation retained an unrestricted recursive position and tested only a marker.

\paragraph{E-06---Validate an uncovered route before accepting a multi-route change.}
\textbf{Type.} \texttt{act}

\textbf{Time point.} \texttt{end}

\textbf{When to refer.} The worker has stopped, and its own trace shows it identified a second route for the delivered behaviour---it inspected or referred to that route---while the committed change leaves that route untouched and no command in the trace ever exercised it.

\textbf{Guidance.} Do not accept yet.
Send the worker back once, naming the route its own trace identified and, where the trace shows it, what that route is missing.
Require the repair and its check to stay within that route: a broad rework at this point can break behaviour that already worked, and the worker no longer has the budget to show that it did not.

\textbf{Why.} In one task, narrow checks missed a failure at the interaction boundary, while in another, an inspected secondary route was left unchanged and unsupported by behavioral evidence; both patterns left a plausible required route unverified at delivery.

\textbf{Reference examples.} (1) \texttt{koota-composite-trait-aspects}: focused checks passed, but a temporary integration exercise exposed a runtime failure that required a committed repair.
(2) \texttt{numba-stencil-boundary-modes}: failed artifacts changed the primary implementation while leaving the previously inspected conversion route untouched.
\end{metaskillcontentbox2}

\end{document}